\documentclass[lettersize,journal]{IEEEtran}
\usepackage{amsmath,amsfonts}
\usepackage{algorithmic}
\usepackage{algorithm}
\usepackage{array}
\usepackage[caption=false,font=normalsize,labelfont=sf,textfont=sf]{subfig}
\usepackage{textcomp}
\usepackage{stfloats}
\usepackage{url}
\usepackage{verbatim}
\usepackage{graphicx}
\usepackage{cite}
\usepackage{generic}
\usepackage{amsmath,amssymb,amsfonts}
\usepackage{algorithmic}
\usepackage{graphicx}
\usepackage{algorithm,algorithmic}
\usepackage{hyperref}
\usepackage{textcomp}
\usepackage{xcolor}
\usepackage[dvipsnames]{xcolor}  
\usepackage{multirow}
\usepackage{graphicx} 
\usepackage[justification=centering]{caption}
\usepackage{amsthm}   
\usepackage{cleveref}
\usepackage{enumerate}
\usepackage{caption} 
\usepackage{booktabs}
\usepackage{tabularx}
\usepackage{float}
\usepackage{amsthm}
\usepackage{pifont}
\usepackage{epstopdf}
\usepackage{xcolor} 
\usepackage{multirow} 
\usepackage{booktabs} 
\usepackage[table,xcdraw]{xcolor}
\usepackage{colortbl}
\usepackage{pifont}
\definecolor{lightblue}{rgb}{0.8, 0.9, 1}
\begin{document}

\title{BARL: Bilateral Alignment in Representation and Label Spaces for Semi-Supervised Volumetric Medical Image Segmentation}

\author{Shujian Gao, Yuan Wang, Zekuan Yu}
\author{Shujian Gao, Yuan Wang, Zekuan Yu
\thanks{This work was partially supported by National Natural Science Foundation of China(82103964).}
\thanks{Shujian Gao and Zekuan Yu are with College of Biomedical and Engineering, Fudan University, Shanghai, 200438, China (e-mails: sjgao24@m.fudan.edu.cn; yzk@fudan.edu.cn).}
\thanks{Yuan Wang is in the ZJU-UIUC Institute, Zhejiang University, Haining, 314400, China (email: yuanwang23@zju.edu.cn).}
\thanks{(Corresponding authors: Zekuan Yu)}}

\markboth{Journal of \LaTeX\ Class Files,~Vol.~14, No.~8, August~2021}%
{Gao \MakeLowercase{\textit{et al.}}: BARL}


\maketitle

\begin{abstract}
Semi-supervised medical image segmentation (SSMIS) seeks to match fully supervised performance while sharply reducing annotation cost. Mainstream SSMIS methods rely on \emph{label-space consistency}, yet they overlook the equally critical \emph{representation-space alignment}. Without harmonizing latent features, models struggle to learn representations that are both discriminative and spatially coherent.
To this end, we introduce \textbf{Bilateral Alignment in Representation and Label spaces (BARL)}, a unified framework that couples two collaborative branches and enforces alignment in both spaces. For label-space alignment, inspired by co-training and multi-scale decoding, we devise \textbf{Dual-Path Regularization (DPR)} and \textbf{Progressively Cognitive Bias Correction (PCBC)} to impose fine-grained cross-branch consistency while mitigating error accumulation from coarse to fine scales. For representation-space alignment, we conduct region-level and lesion-instance matching between branches, explicitly capturing the fragmented, complex pathological patterns common in medical imagery.
Extensive experiments on four public benchmarks and a proprietary CBCT dataset demonstrate that BARL consistently surpasses state-of-the-art SSMIS methods. Ablative studies further validate the contribution of each component. Code will be released soon.
\end{abstract}

\begin{IEEEkeywords}
Semi-Supervised Medical Image Segmentation, Consistency Regularization, Representation Learning
\end{IEEEkeywords}

\section{Introduction}
\label{sec:introduction}
\begin{figure}[t] 
    \centering
    \includegraphics[width=8.0cm]{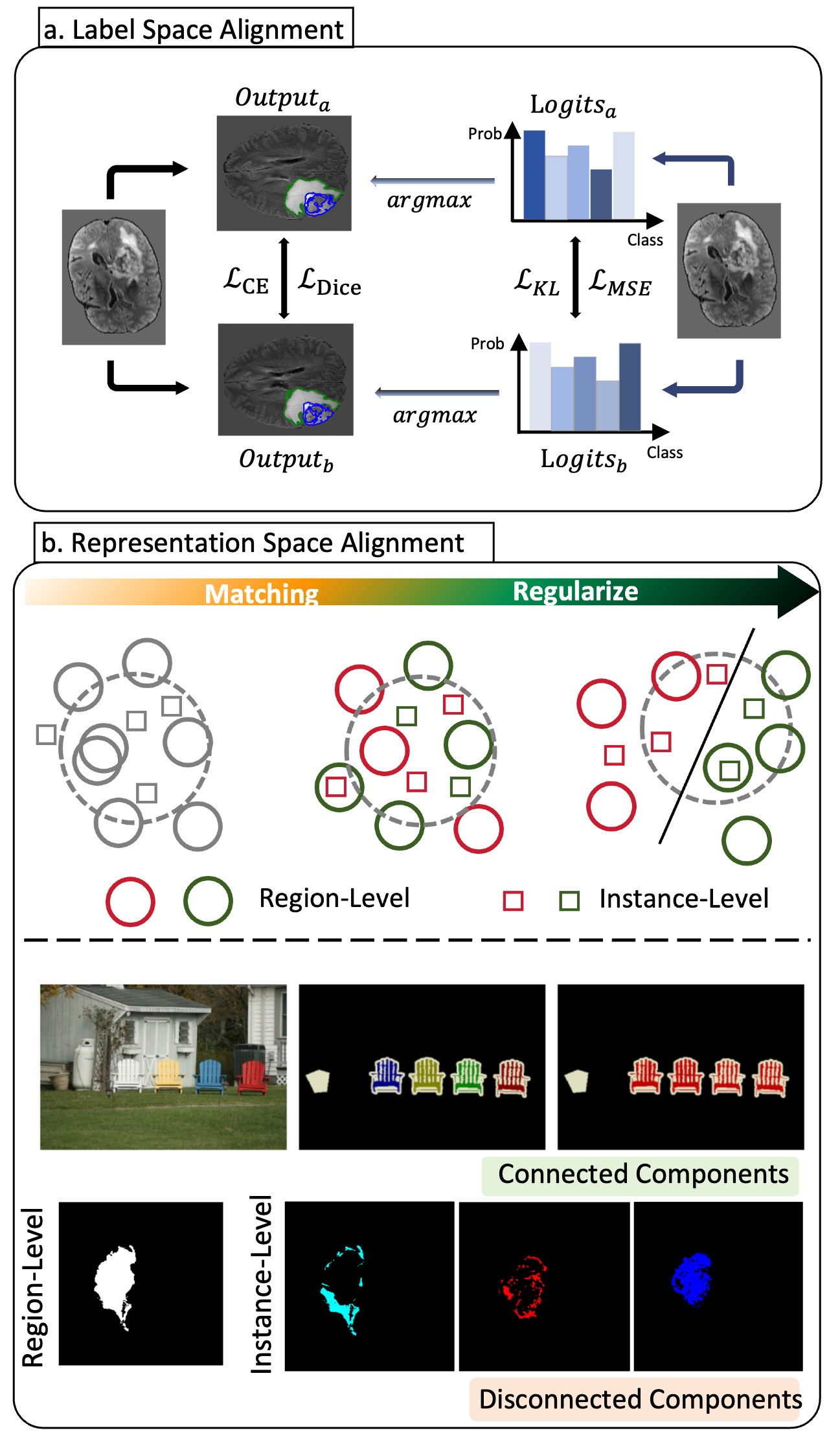}
    \captionsetup{justification=justified, singlelinecheck=false} 
    \caption{\textbf{Typical alignment protocols and lesion fragmentation statistics.}  
\textbf{(a) Label space.} Soft logits are aligned by Kullback--Leibler or mean-squared error, whereas hard pseudo-labels (via \textit{arg\,max}) are aligned by Dice or cross-entropy.  
\textbf{(b) Representation space.} We locate class-specific feature vectors, then enforce alignment to widen inter-class margins while tightening intra-class clusters. 
\textbf{Fragmentation in Medical Images.}
\emph{Top}: A natural scene with its category-level mask; pixels of each class coalesce into a single, compact connected component. 
\emph{Bottom}: A medical mask. Left—foreground (\textit{white}) vs.\ background (\textit{black}); right—lesions encoded by colour, where voxels of the same class split into multiple disconnected fragments.}
    \label{fig:figure1} 
\end{figure}
\IEEEPARstart{M}{edical} image segmentation is widely regarded as a cornerstone of modern computer-aided diagnosis\cite{intro1, intro2}, intra-operative navigation \cite{intro3}, and quantitative image understanding \cite{intro4}.  Recent breakthroughs in data-driven deep learning have elevated performance across diverse clinical modalities, such as ultrasound\cite{intro_ultra}, magnetic resonance imaging\cite{intro_mri}, and computed tomography\cite{intro_ct}, and have yielded substantial gains in lesion delineation \cite{intro_lesion} and therapeutic planning \cite{intro_plan}.  Nevertheless, the severe scarcity of pixel-wise annotations, compounded by the high cost and subjectivity of manual contouring, continues to constrain segmentation accuracy \cite{intro_anno}.  Under such extreme supervision deficits, networks can only access partial annotated data, inevitably learning biased or incomplete representations that propagate systematic errors at inference time \cite{intro_bias}.  Hence, developing learning algorithms capable of learning rich, unbiased knowledge from datasets in which merely a \emph{small fraction} of images are annotated has become a fundamental prerequisite for robust and reliable medical image segmentation \cite{intro5}.

SSMIS has emerged as a compelling paradigm for alleviating above issues. Recent studies pursue this goal through several complementary strategies, including consistency-based \emph{alignment} that enforces prediction invariance under stochastic perturbations \cite{cct} and geometric transformations \cite{dtc, cct2}, \emph{co-training} where dual subnetworks exchange pseudo-labels to achieve mutual supervision\,\cite{cps}, \emph{adversarial learning} that employs discriminators to align feature and prediction distributions between labelled and unlabelled data\,\cite{mtans}, \emph{contrastive learning} which drives latent embeddings of identical anatomical structures to cluster while repelling dissimilar ones \cite{contras1, contras2}, and iterative \emph{self-training} schemes that refine pseudo-labels to expand the effective supervision set\,\cite{self-training}. Collectively, these advances substantially reduce the reliance on exhaustive manual delineation, lowering operational costs and expediting the production of reliable segmentation masks for medical images across different modalities.

Among above-mentioned strategies, consistency regularization has become the \emph{de-facto} principle of SSMIS \cite{consist}, predicated on the smoothness assumption \cite{smooth} that spatially \cite{fixmatch} or photometrically \cite{uda} perturbed inputs should yield consistent predictions under either identical \cite{remixmatch, mixmatch} or heterogeneous \cite{sun2024semi} networks.  Within this framework, \textit{alignment} can be divided into two complementary categories, as depicted in Figure \ref{fig:figure1}.  \textbf{Label-space alignment} directly supervises results in output heads: \emph{soft} logits are regularized with Kullback–Leibler divergence \cite{dtc} or mean-squared error (MSE) \cite{mt}, whereas \emph{hard} arg-max pseudo-labels are compared with Dice \cite{ijcnn} or cross-entropy (CE) losses \cite{mct, pseudoseg}.  In parallel, \textbf{representation-space alignment} constrains intermediate feature embeddings so that the representation features between categories are aligned, typically via positive-only contrastive \cite{posionly}, cosine \cite{cross}, or center-loss \cite{centerloss} objectives.

Existing consistency schemes fail to comprehensively enforce \emph{dual-space} alignment (simultaneous regulation of both label and representation spaces) and omit the highly fragmented distribution of lesions in medical images.
Most works enforce consistency only in the output label space, for instance, by matching predictions for different augmentations of an image \cite{fixmatch, mixmatch, mctv1} or by aligning class posterior distributions \cite{remixmatch}. While effective, these approaches often neglect the underlying structure of the learned intermediate feature representations. We argue that merely aligning outputs can be insufficient, especially when feature manifolds are misaligned \cite{crmatch} or when class boundaries are complex \cite{edgeaware} in the representation space. Although \cite{information,posionly} introduce dual-space regularization, they are designed for natural images and do not account for the intricate pathological characteristics present in medical images, \emph{i.e.}, lesions of the same class often appear fragmented, as illustrated in Figure~\ref{fig:figure1}(b) and discussed in Section~\ref{sec:lesion-level}. To fully exploit consistency regularization, we propose BARL, which concurrently aligns data distributions in both the label and representation spaces, unifying these two complementary paradigms into a single optimization objective. Our contributions are four-fold:

\begin{itemize}

\item \textbf{Problem formulation.}  
We formulate semi-supervised segmentation as a dual–space constraint optimisation problem, simultaneously regularizing feature distributions in the \emph{representation space} and the \emph{label space}.

\item \textbf{Dual–space regularization.}  
In the label space, DPR and PCBC respectively leverage multi-level decoder semantics and rectify cross-branch discrepancies.
In the representation space, we align features at both a coarse \emph{region} level and a fine-grained \emph{instance} level, capturing the complex pathological characteristics and fostering inter-branch latent features consistency.

\item \textbf{Comprehensive evaluation.}  
Experiments on four public 3-D medical segmentation benchmarks and a private dental CBCT dataset under multiple labelled–to–unlabelled ratios show that BARL consistently surpasses state-of-the-art counterparts, yielding promising gains. A suite of ablation studies further confirm the individual contribution of each proposed component. 

\item \textbf{In-depth analysis.}  
Beyond basic experiments, we conduct a series of \emph{exploratory investigations}, distinct from comparative baseline approaches. These include comparisons of consistency-regularization techniques, analyses of representation spaces and their dimensionality, and examinations of alternative semi-supervised segmentation frameworks. Our goal is to provide additional theoretical insights and empirical evidence for semi-supervised segmentation research.
\end{itemize}

\section{Related Work}
\subsection{Medical Image Segmentation}
Over the past decade, deep learning has emerged as the cornerstone of medical image segmentation, driven by its ability to deliver high-throughput processing, full automation, and near-expert accuracy \cite{intro1, intro3, intro5}. Current methods can be broadly categorized into three architectural families: (i) Convolutional Neural Networks (CNNs) \cite{unet, attentionunet, vnet}, (ii) Vision Transformers (ViTs) \cite{segvit, swin}, and (iii) Hybrid models combining CNNs and ViTs \cite{cnntrans}. While ViTs, with their self-attention mechanisms, excel at capturing long-range dependencies, CNNs remain the dominant choice in clinical settings. This is largely due to their favorable trade-offs: faster inference, lower memory usage, and better compatibility with standard hardware. In this work, we benchmark multiple backbone architectures within our BARL framework, offering empirical comparisons that shed light on the relative strengths of convolution and self-attention under real-world medical constraints.

Methodologically, medical image segmentation is typically conceptualized at two distinct spatial granularities: \emph{(a)}~two-dimensional pixel-wise delineation of individual image slices \cite{centerloss}, and \emph{(b)}~three-dimensional (3-D) voxel-wise delineation of volumetric data \cite{mt, effective}. The latter presents substantially greater challenges, primarily attributable to characteristics inherent in volumetric scans \cite{intro2, intro3}, such as anisotropic resolution, intricate anatomical topologies, and pronounced inter-slice dependencies. Consequently, achieving accurate yet computationally efficient 3-D segmentation persists as an active and critical research frontier \cite{ra3}. In this paper, we focus on the fine-grained segmentation of 3D medical data under conditions of scarce annotations, establishing new state-of-the-art for semi-supervised voxel segmentation.


\subsection{Semi-Supervised Image Segmentation}

Semi-supervised image segmentation (SSIS) algorithms aim to train models using a small amount of labeled data and a large amount of unlabeled data \cite{rb2}. Compared to approaches that use only limited labeled data or no labeled data at all, SSIS offers a more efficient and practical solution. Here, we detail some fundamental strategies. Pseudo-labeling \cite{pseudoseg, self-training, pseudoalign} is an early semi-supervised learning algorithm that improves model performance through iterative inference and re-training on unlabeled data. The focus of this paradigm is on removing unreliable pseudo-labels through fixed threshold filtering \cite{contras2}, dynamic confidence filtering \cite{centerloss}, or auxiliary network filtering \cite{auxiliary}, and actively improving pseudo-label quality through label correction \cite{iplc} and bias elimination \cite{debias}.

Consistency regularization methods are based on the smoothness assumption \cite{smooth} and leverage unlabeled data to learn more robust feature representations \cite{r1}. Common types of perturbations include: data-level perturbations such as noise injection \cite{cct, crossmatch}, weak-to-strong augmentation \cite{fixmatch, r6}, color jitter, cutout \cite{cutout}, cutmix \cite{cutmix}, classmix \cite{classmix}, etc.; model-level perturbations involving homogeneous \cite{cps, mt} or heterogeneous model \cite{mct} architectures, such as single encoder-multiple decoder architectures \cite{cct} or Mean Teacher (MT) architectures \cite{mt}; and feature-level perturbations including feature Dropout \cite{crossmatch} or feature noise \cite{noise}. Furthermore, adversarial-based methods \cite{mtans}, co-training \cite{cps, cct2}, multi-view learning \cite{uamc}, and entropy minimization \cite{posionly} have also played significant roles in SSIS.

Beyond basic algorithms, semi-supervised learning frameworks also hold considerable importance. Fig. \ref{fig:figure3} illustrates four prevailing mainstream architectures \cite{cps, fixmatch, mctv1, uamt}. We argue that the selection of an appropriate architecture, and the subsequent integration of consistency constraints with corresponding data augmentation techniques, constitutes the core methodology of SSIS. In this paper, we further investigate the efficacy of the BARL algorithm across these diverse frameworks, thereby providing empirical validation for the significant role of semi-supervised learning frameworks.


\subsection{Consistency Regularization in SSIS}
Consistency regularization is currently considered a mainstream approach in semi-supervised learning algorithms. Its core idea is that model predictions for unlabeled data should remain consistent after applying different perturbations, such as data augmentation or noise. This mechanism allows the model to learn the intrinsic structure and robustness of the data from unlabeled samples, thereby improving generalization ability. Existing methods primarily enforce alignment within the label space, including: $\Pi$ Model, Temporal Ensembling, Mean Teacher \cite{mt}, MixMatch \cite{mixmatch}, ReMixMatch \cite{remixmatch}, and FixMatch \cite{fixmatch}.

However, merely enforcing consistency in the label space may not be sufficient to fully leverage the potential of unlabeled data, especially when learning complex visual representations. CR-Match \cite{crmatch} puts forth a Feature Distance Loss aimed at regularizing the representation distribution. A limitation, however, is its predominant focus on the global latent distribution, failing to achieve class-wise and region-wise alignment. \cite{posionly} utilizes a positive-only learning scheme to align same-class features within the MT framework, it presents two significant limitations. First, its reliance on a memory bank introduces considerable computational and memory overhead. Second, it does not account for the characteristically discrete and often fragmented distribution of lesions in medical imaging. Therefore, we propose BARL, which enforces data consistency across both a fine-grained representation space and multi-perspective label space. By aligning features in the representation space, BARL enhances class compactness and inter-lesion separability, thereby boosting generalization to complex anatomical structures such as fragmented lesions.

\section{Method}
\subsection{Overview and Preliminary}
\begin{figure*}[t!]  
    \centering
    \includegraphics[width=\linewidth]{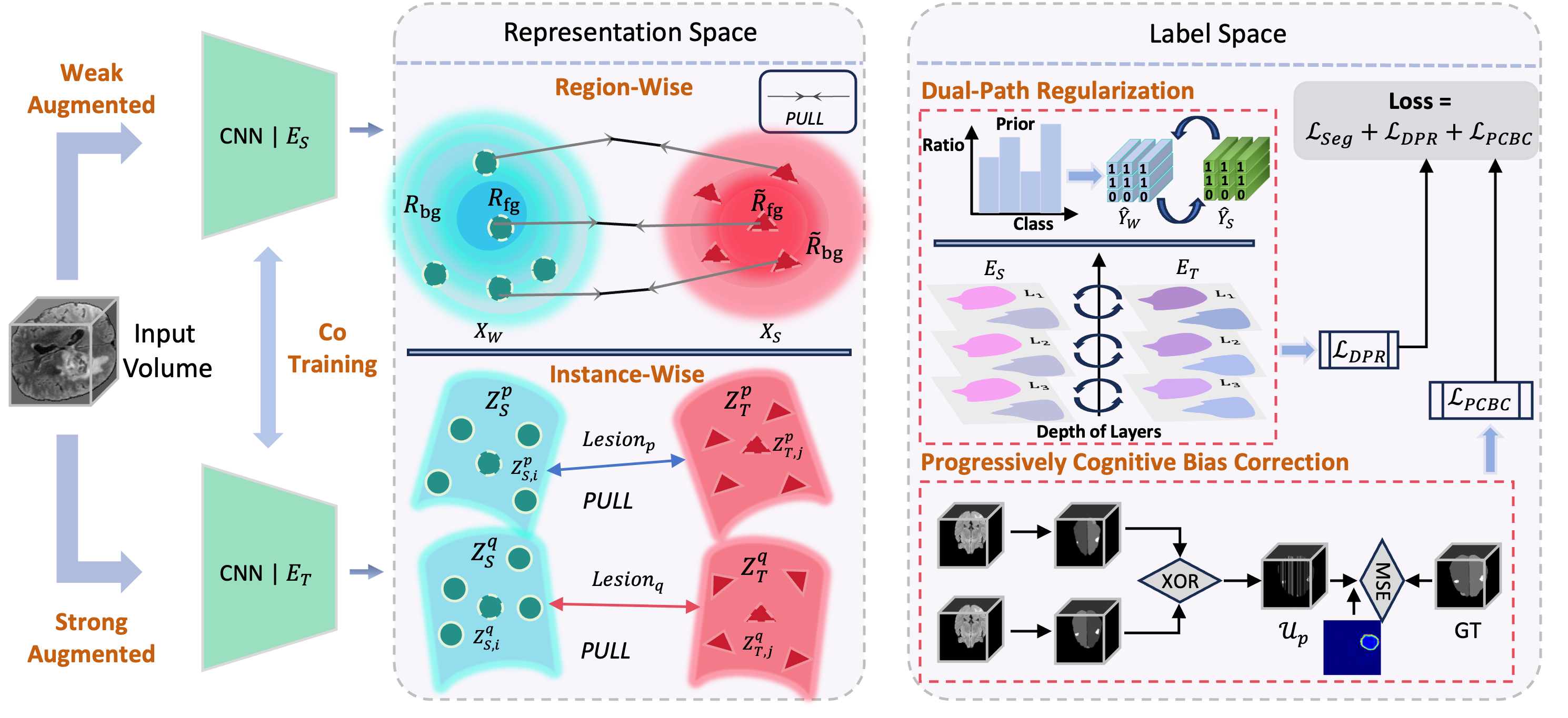}
    \captionsetup{justification=justified, singlelinecheck=false} 
    \caption{Overview of our proposed BARL strategy, built upon a co-training architecture. The strategy enforces constraints in two dimensions: the representation space, where region-level and instance-level feature vectors are obtained and consistency operations are applied; and the label space, where the proposed DPR module constrains outputs among multi-layer decoders and the PCBC module corrects inconsistent results under the co-training framework.}
    \label{fig:figure2} 
\end{figure*}
Given a labeled set $\mathcal{D}_L = \left\{(x_i^l, y_i^l)\right\}_{i=1}^{N_l}$ and a much larger unlabeled set $\mathcal{D}_U = \left\{x_i^u, y_i^u\right\}_{i=1}^{N_u}$ where $N_u \gg N_l$ and $x_i \in \mathbb{R}^{C \times D \times H \times W}$ represents a 3D volumetric image with $C$ channels and spatial dimensions $D$, $H$, and $W$, our objective is to fully exploit the information contained in the unlabeled data under the guidance of a limited amount of annotated samples, thereby achieving superior segmentation performance \cite{intro5}.

To achieve this, we propose the BARL framework, as illustrated in Fig. \ref{fig:figure2}. BARL is rooted in the classical \emph{co‑training} paradigm \cite{cps} and employs two parameter‑independent models, denoted as $E_S$ and $E_T$, which enables better feature learning capabilities within limited labeled data.

In general, the BARL framework can be outlined into two parts:

\pmb{Part 1}~\ref{Representation Space Alignment}: To enforce tighter structural constraints inside the latent space, we perform alignment operations separately on region-level and lesion-instance-level features.

\pmb{Part 2}~\ref{sec:label space alignment}: To unleash the potential of label‑space alignment in the context of the co‑training architecture, we introduce DPR and PCBC modules that constrain and refine the feature distribution in the label space from multiple perspectives.
\subsection{Representation Space Alignment}
\label{Representation Space Alignment}
Previous SSIS methods have predominantly concentrated on enforcing constraints in the \emph{label space}, for example, by matching student–teacher predictions \cite{uamt} or refining pseudo‑labels \cite{iplc, debias}. In doing so, they have largely overlooked the equally important goal of aligning the \emph{representation space}. Without an explicit mechanism that draws semantically similar features closer together, a network can satisfy a label‑space consistency loss and yet still learn a disordered latent space, resulting in ambiguous or fragmented masks \cite{crmatch}. As shown in Fig. \ref{fig:figure2}, we regularize the representation distributions at both the region‑level and lesion‑instance-level.

\subsubsection{Region-Level}
To enforce representation feature consistency, we introduce a \textbf{Region-Wise} alignment mechanism, as illustrated in Figure~\ref{fig:figure2}. Specifically, given an input image, we generate a weakly-augmented view $X_w$ and a strongly-augmented view $X_s$. $E_S$ processes the weakly-augmented view $X_w$ to produce a high-level feature map $\mathbf{f}_S = E_S(X_w)$, while $E_T$ processes the strongly-augmented view $X_s$ to yield $\mathbf{f}_T = E_T(X_s)$. The models also produce segmentation probability maps $P_S$ and $P_T$, respectively.

The set $\mathcal{C}_{\mathrm{region}}$ includes all foreground (fg) and background (bg) classes. For each category $c \in \mathcal{C}_{\mathrm{region}}$, we obtain binary segmentation masks $M_S^c$ and $M_T^c$ by applying the $\arg\max$ operation to $P_S$ and $P_T$.

We then extract region-level representations based on these masks. For each class $c$, the prototype vector from the encoder $E_S$, denoted as $\mathbf{R}_c$ (e.g., $\mathbf{R}_{\mathrm{fg}}$, $\mathbf{R}_{\mathrm{bg}}$ in fig.2), is computed as the mean of feature vectors $\mathbf{f}_S(p)$ over all spatial locations $p$ identified by the mask $M_T^c$:
\begin{equation}
    \mathbf{R}_c = \frac{1}{|M_S^c|} \sum_{p \in M_S^c} \mathbf{f}_T(p)
    \label{eq:proto_r}
\end{equation}
where $p$ indexes the spatial locations, and $|M_S^c|$ is the number of pixels in region $M_T^c$.

Similarly, the corresponding embedding from the encoder $E_T$, denoted as $\mathbf{\tilde{R}}_c$ (e.g., $\mathbf{\tilde{R}}_{\mathrm{fg}}$, $\mathbf{\tilde{R}}_{\mathrm{bg}}$), is computed using the features $\mathbf{f}_T$ and mask $M_T^c$:
\begin{equation}
    \mathbf{\tilde{R}}_c = \frac{1}{|M_T^c|} \sum_{p \in M_T^c} \mathbf{f}_S(p)
    \label{eq:proto_r_tilde}
\end{equation}

The goal of the alignment is to \emph{PULL} the embedding $\mathbf{\tilde{R}}_c$ towards the embedding $\mathbf{R}_c$. This is achieved by minimizing the cosine distance between them. The alignment loss is:
\begin{equation}
    L_{\mathrm{region}} = 1 - \frac{\mathbf{R}_c \cdot \mathbf{\tilde{R}}_c}{\|\mathbf{R}_c\|_2 \|\mathbf{\tilde{R}}_c\|_2}, c \in \mathcal{C}_{\mathrm{region}}
    \label{eq:loss_align_c}
\end{equation}
where $\|\cdot\|_2$ denotes the L2 norm. 

This loss enforces region-level representation consistency between the outputs of the two models. By encouraging the representations from the weakly-augmented input to match those from the strongly-augmented input, the model learns robust features that are invariant to the strength of data augmentation.

\subsubsection{Instance-Level}
\label{sec:lesion-level}
As shown in Fig.~\ref{fig:figure1}(b), the anatomical structure of medical imaging data inherently comprises not only foreground-background differentiation but also different lesion categories. However, these critical lesions frequently exhibit fragmented spatial distribution patterns, lesion instances within the same category often appear spatially distributed as discrete clusters. Owing to this inherent fragmentation, directly computing a lesion-level prototype is ill-posed and can introduce an prototype-specific shift. To address this challenge, we develop a fine-grained \emph{instance-level} alignment framework that ensures feature consistency between corresponding lesion instances processed by the $E_S$ and $E_T$. This mechanism is specifically designed to promote anatomical coherence within the feature space while preserving pathological characteristics across different images.

Inspired by 3D connected-component analysis~\cite{cc3d}, we extract individual lesion instances from the binary masks. To ensure stability, we perform this operation exclusively on the output of the more stable encoder, $E_T$. Given the $E_T$'s binary mask $M_T^c$ for a lesion category $c \in \mathcal{C}_{\mathrm{lesion}}$, we apply a 3D connected-component labeling operator:
\begin{equation}
    \{M_j^c\}_{j=1}^{N^c} = \text{ConnComp3D}(M_T^c),
    \label{eq:conncomp}
\end{equation}
where $\text{ConnComp3D}(\cdot)$ is the extraction operator and $\{M_j^c\}$ is the resulting set of individual lesion instance masks. Each mask $M_j^c$ identifies a candidate lesion instance as a contiguous region of voxels. To mitigate noise, we filter out small, insignificant components by enforcing a minimum volume threshold $\tau_{\text{vol}}$:
\begin{equation}
    |M_j^c| \geq \tau_{\text{vol}},
    \label{eq:filter_volume}
\end{equation}
where $|M_j^c|$ denotes the number of voxels in the instance mask. The set of remaining components represents the definitive lesion instances for alignment.

For each identified lesion instance $M_j^c$, we compute a corresponding pair of prototype vectors by pooling features from both $E_S$ and $E_T$ within that same instance mask. The prototype $\mathbf{z}_{S,j}^c \in \mathbb{R}^D$ is the average of its feature vectors over the instance:
\begin{equation}
    \mathbf{z}_{S,j}^c = \frac{1}{|M_j^c|} \sum_{p \in M_j^c} \mathbf{f}_S(p).
    \label{eq:proto_student}
\end{equation}

Similarly, the prototype $\mathbf{z}_{T,j}^c$ is computed using the feature map $\mathbf{f}_T$:
\begin{equation}
    \mathbf{z}_{T,j}^c = \frac{1}{|M_j^c|} \sum_{p \in M_j^c} \mathbf{f}_T(p).
    \label{eq:proto_teacher}
\end{equation}

This process yields a set of directly corresponding prototype pairs $(\mathbf{z}_{S,j}^c, \mathbf{z}_{T,j}^c)$ for each lesion instance $j$. Since each instance provides a natural one-to-one correspondence, we can directly define an instance-level alignment loss to maximize the similarity between prototype pairs. For each class $c \in \mathcal{C}_{\mathrm{lesion}}$, the loss is formulated as the average cosine distance over all $N^c$ detected instances:
\begin{equation}
    \mathcal{L}_{\text{instance}} = \frac{1}{N^c} \sum_{j=1}^{N^c} \left(1 - \frac{\mathbf{z}_{S,j}^c \cdot \mathbf{z}_{T,j}^c}{\|\mathbf{z}_{S,j}^c\|_2 \|\mathbf{z}_{T,j}^c\|_2}\right), c \in \mathcal{C}_{\mathrm{lesion}}
    \label{eq:align_loss}
\end{equation}

This module compels $E_S$ and $E_T$ to yield congruent embeddings for the same physical lesion, which facilitates the learning of discriminative, instance-level features by enforcing direct feature alignment between the two models.

\subsection{Label Space Alignment}
\label{sec:label space alignment}
The current mainstream semi-supervised alignment algorithms primarily impose constraints on soft logits \cite{dtc, mt, crossmatch} or hard pseudo-labels \cite{ijcnn, mct, pseudoseg} within the label space. However, these methods fail to fully exploit the unique characteristics of co-training architectures, particularly the alignment between multi-level decoders and the inherent divergence in dual-branch outputs. To address this limitation, we propose the DPR module and the PCBC module. 

\subsubsection{Dual-Path Regularization}

The model backbone employed in this study comprises an encoder and a multi-layer decoder. Within the co-training framework, existing approaches that solely align the segmentation output heads of $E_S$ and $E_T$ fail to account for the hierarchical characteristics of the decoder architecture. Furthermore, we identify two critical limitations in conventional constraint strategies: (1) Applying constraints solely on probability logits \cite{mctv1} inadequately captures structural information due to their uncalibrated nature, and (2) Relying exclusively on hard pseudo-labels \cite{cps} introduces noise propagation risks from erroneous predictions. Hence, we propose a comprehensive layer-wise alignment mechanism that systematically integrates constraint operations across all decoder layers.

Both $E_S$ and $E_T$ decode representations at a hierarchy of scales. Alongside the main segmentation map at full resolution, we attach three auxiliary output heads to intermediate layers, yielding predictions at coarser resolutions. We denote these outputs as $P_S^{(3)}, P_S^{(2)}, P_S^{(1)}, P_S^{(0)}$ for $E_S$ (with $P_S^{(3)}$ being the final full-resolution prediction and $P_S^{(0)}$ the coarsest auxiliary prediction). Similarly, $P_T^{(k)}$ for $k=0,1,2,3$ are the outputs for $E_T$. While the focus of feature representation varies across different layers, the salient characteristics of the primary lesion region are robustly extracted throughout \cite{erdunet}. Such a multi-scale configuration is instrumental for enforcing a detailed, hierarchical consistency between $E_S$ and $E_T$.

\noindent\textbf{Dual-Path Consistency Loss:} We impose consistency between $E_S$ and $E_T$ predictions at all scales via two complementary loss terms. 

\textit{(i) Distributional Consistency Loss ($\mathcal{L}_{\text{distr}}$):} 
To encourage $E_S$ and $E_T$ to produce similarly shaped probability distributions, we penalize the MSE between their softened predictions, which aligns the overall confidence landscape at each scale:
\begin{equation}
    \mathcal{L}_{\text{distr}} = \frac{1}{4}\sum_{k=0}^{3} \left\| \text{sPL}(P_S^{(k)}, T) - \text{sPL}(P_T^{(k)}, T) \right\|_2^2,
\end{equation}
where $\text{sPL}(\cdot, T)$ denotes the softening function with temperature $T$.

\textit{(ii) Deep Cross Pseudo Supervision Loss ($\mathcal{L}_{\text{CPS}}$):} 
Inspired by CPS\cite{cps}, we employ a cross-supervision mechanism where each model learns from the other's confident predictions. We generate one-hot pseudo-labels $\hat{P}^{(k)}$ from each model's output. $E_S$ is then supervised by the $E_T$'s pseudo-labels, and vice versa, using a standard CE loss:
\begin{equation}
    \mathcal{L}_{\text{CPS}} = \frac{1}{4} \sum_{k=0}^{3} \left[ \mathcal{L}_{\text{CE}}\left(P_S^{(k)}, \hat{P}_T^{(k)}\right) + \mathcal{L}_{\text{CE}}\left(P_T^{(k)}, \hat{P}_S^{(k)}\right) \right],
\end{equation}

\noindent\textbf{Information Maximization Loss:} Besides enforcing consistency between the dual paths, we impose an \emph{information maximization} regularization \cite{imloss} on the predicted label distributions. This consists of two parts aimed at achieving confident yet well-distributed predictions. To reduce predictive uncertainty on unlabeled data, we employ an entropy minimization loss, $\mathcal{L}_{\text{ent}}$.
\begin{equation}
    \mathcal{L}_{\text{ent}} = \mathbb{E}_{p \sim P_S} \left[ -\sum_{c=1}^{C} p_c \log p_c \right],
    \label{eq:ent_loss}
\end{equation}
where $p$ is the $C$-dimensional probability vector for a single pixel in the model's output map $P_S$. Minimizing this loss pushes each prediction vector $p$ towards a one-hot distribution, thereby increasing prediction confidence.

Second, to prevent the model from collapsing to trivial solutions (e.g., predicting only the background), we introduce a regularization term. This term aligns the model's average predicted class distribution, $\bar{\mathbf{p}}$, with a predefined class prior, $\mathbf{q}$, using the modified KL divergence.

We define the model's average prediction $\bar{\mathbf{p}} = (\bar{p}_1, \dots, \bar{p}_C)$, where each component is the mean probability for that class over all $N$ pixels:
\begin{equation}
    \bar{p}_c = \frac{1}{N}\sum_{i=1}^{N} p_{i}^{c}.
\end{equation}

The target prior $\mathbf{q} = (q_1, \dots, q_C)$ is derived from the empirical class frequencies observed in the labeled training set.
Then, the prior-matching loss is the KL divergence from the target prior $\mathbf{q}$ to the predicted distribution $\bar{\mathbf{p}}$:
\begin{equation}
    \mathcal{L}_{\text{KL}} = \sum_{c=1}^{C} \bar{p}_c \log\left(\frac{\bar{p}_c}{q_c}\right).
\end{equation}

Minimizing $\mathcal{L}_{\text{KL}}$ encourages the model to produce predictions that respect the overall class proportions, promoting diversity and counteracting predictive collapse.

\noindent\textbf{Total Loss:} The final regularization loss, $\mathcal{L}_{\text{DPR}}$, combines the dual-path consistency terms with a weighted Information Maximization (IM) term. The IM loss itself is composed of the $\mathcal{L}_{\text{ent}}$ and the class-prior matching loss ($\mathcal{L}_{\text{KL}}$). Formulated as:
\begin{equation}
    \mathcal{L}_{\text{DPR}} = \mathcal{L}_{\text{distr}} + \mathcal{L}_{\text{DeepCPS}} + \left( \lambda_{\text{ent}} \mathcal{L}_{\text{ent}} + \lambda_{\text{KL}} \mathcal{L}_{\text{KL}} \right),
\end{equation}
where $\lambda_{\text{ent}}$ and $\lambda_{\text{KL}}$ are hyperparameters that balance the components of the IM loss. In our implementation, we set their values to $\lambda_{\text{ent}} = 0.5$ and $\lambda_{\text{KL}} = 0.1$.

\subsubsection{Progressively Cognitive Bias Correction}

\begin{figure}[t!]  
    \centering
    \includegraphics[width=0.9\linewidth]{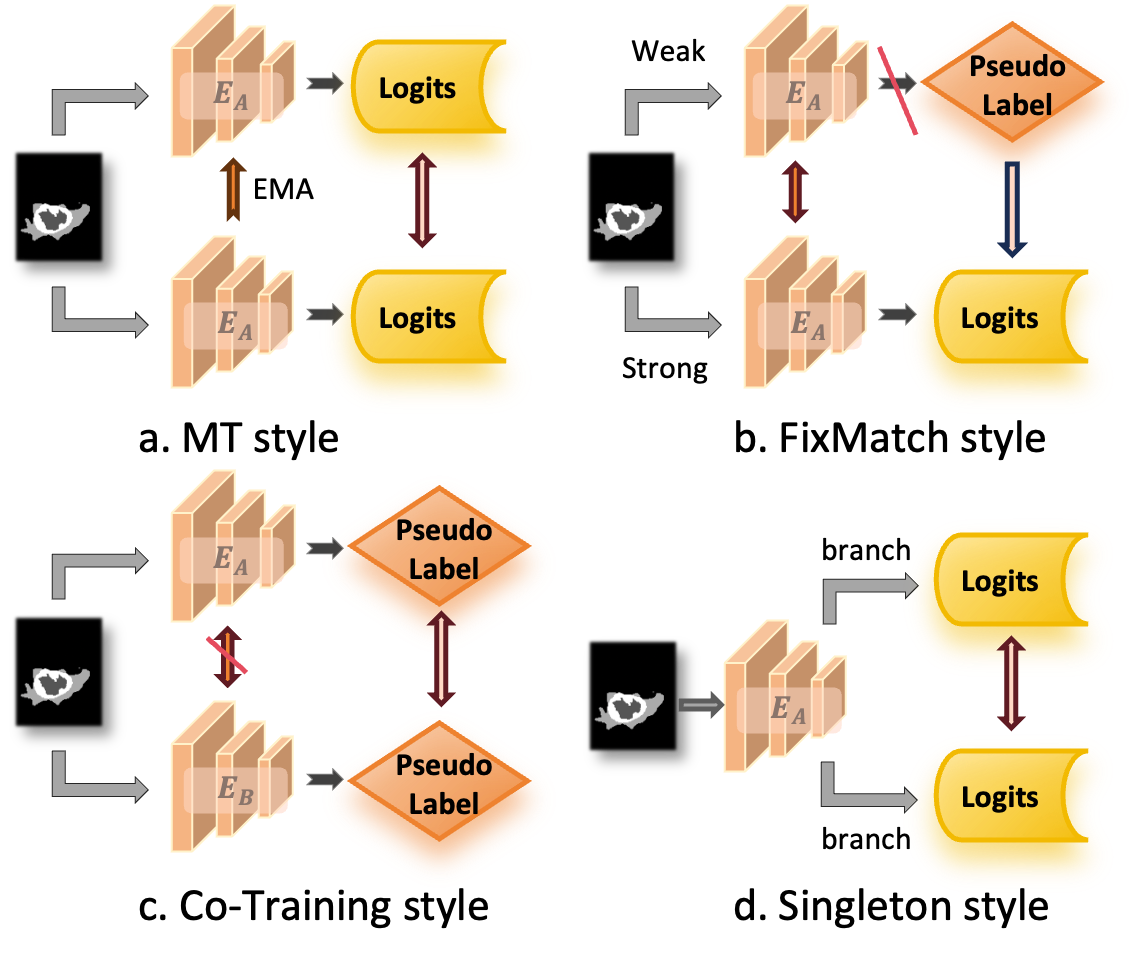}
    \caption{\raggedright A summary of mainstream semi-supervised learning architectures, including Mean Teacher style, FixMatch-style consistency regularization, Co-Training with mutual pseudo labeling, and Singleton-style frameworks.}
    \label{fig:figure3} 
\end{figure}

Not all regions of labeled images are equally challenging; $E_S$ and $E_T$ may confidently agree in clear regions while disagreeing in ambiguous ones \cite{mcf}. The key to enhancing model performance lies in optimizing the regions of discrepancy\cite{erdunet}. To harness this disagreement as a proxy for model uncertainty, we introduce an uncertainty-guided loss that adaptively focuses the cognitive bias correction on these contentious areas.

To leverage inter-model divergence as a proxy for pixel-wise uncertainty, we define a continuous uncertainty weight, $\mathcal{U}_p$. Unlike binary hard-masking approaches based on final predictions \cite{mcf}, our soft weight quantifies the \emph{degree} of disagreement using the L1 distance between the $E_S$ and $E_T$ probability distributions:
\begin{equation}
\mathcal{U}_p = \left\|P_{S,p} - P_{T,p}\right\|_1,
\end{equation}
where $P_{S,p}, P_{T,p} \in \mathbb{R}^C$ are the respective probability vectors over $C$ classes.

This uncertainty weight then modulates a MSE loss, yielding our final uncertainty-guided loss function. This loss is an uncertainty-weighted average MSE between the models' predictions and the one-hot ground truth label $y^l$:
\begin{equation}
\mathcal{L}_{\mathrm{PCBC}} = \frac{\sum_{p\in\Omega} \mathcal{U}_p \left( \bigl\|P_{S,p} - y^l_p\bigr\|_2^2 + \bigl\|P_{T,p} - y^l_p\bigr\|_2^2 \right)}{\sum_{p\in\Omega} \mathcal{U}_p + \epsilon},
\end{equation}
where $y^l_p \in \{0,1\}^C$ is the one-hot ground truth label at pixel $p$ and $\epsilon$ is a small constant for numerical stability.

This formulation ensures that pixels with higher uncertainty incur a proportionally larger penalty. By compelling both models to specifically resolve their most significant disagreements and align with the ground truth in these ambiguous regions, our uncertainty-guided loss effectively refines predictions where they matter most, ultimately enhancing the overall segmentation accuracy of the co-training framework.
\subsubsection{Segmentation Loss}
For labeled data, the segmentation process is guided by a hybrid supervised loss function, combining CE and Dice losses to ensure both pixel-level accuracy and spatial overlap quality. Given the network's prediction $P$ and the GT $y_i^l$, the overall supervised loss is formulated as:
\begin{equation}
    \mathcal{L}_{seg}(P, y_i^l) = \mathcal{L}_{CE}(P, y_i^l)) + \mathcal{L}_{Dice}(P, y_i^l))
    \label{eq:sup_loss_combined}
\end{equation}

\subsection{Overall Learning Objective}
In the end-to-end training, the total loss is shown below:
\begin{equation}
\mathcal{L}_{\text{s}} = 0.1 \times (\mathcal{L}_\text{region} + \mathcal{L}_\text{instance}) + \mathcal{L}_\text{DPR} + \mathcal{L}_\text{PCBC} + \mathcal{L}_\text{seg} .
\end{equation}

\section{EXPERIMENTS}
\subsection{Experimental Setup}
\subsubsection{Datasets}

\begin{table*}[htbp]
\centering
\caption{Quantitative evaluation of different baselines on BraTs 2021, 2020, BraTs 2023 MEN, and CBCT Tooth datasets under 10\% and 20\% label ratio. \textcolor{red}{Red}-colored and \textcolor{blue}{Blue}-colored values correspond to the best and 2nd best performing model, respectively.}
\resizebox{\textwidth}{!}{ 
\begin{tabular}{l|l|cccc|cccc|cccc|cccc}
\toprule
\multicolumn{1}{c|}{\multirow{2}{*}{\textbf{Ratio}}} & \multicolumn{1}{c|}{\multirow{2}{*}{\textbf{Method}}} & \multicolumn{4}{c|}{BraTs2020} & \multicolumn{4}{c|}{BraTs2021} & \multicolumn{4}{c|}{BraTs2023 MEN} & \multicolumn{4}{c}{Tooth CBCT} \\
\multicolumn{1}{c|}{} & \multicolumn{1}{c|}{} & \textit{Dice $\uparrow$} & \textit{HD $\downarrow$} & \textit{ASD $\downarrow$} & \textit{Jaccard $\uparrow$} & \textit{Dice $\uparrow$} & \textit{HD $\downarrow$} & \textit{ASD $\downarrow$} & \textit{Jaccard $\uparrow$} & \textit{Dice $\uparrow$} & \textit{HD $\downarrow$} & \textit{ASD $\downarrow$} & \textit{Jaccard $\uparrow$} & \textit{Dice $\uparrow$} & \textit{HD $\downarrow$} & \textit{ASD $\downarrow$} & \textit{Jaccard $\uparrow$} \\
\midrule
\multirow{14}{*}{10\%} & CPS \textcolor{gray}{\scriptsize [CVPR 2021]} & 0.8262 & \textcolor[HTML]{0000FF}{9.2266} & \textcolor[HTML]{0000FF}{2.9447} & \textcolor[HTML]{000000}{0.7303} & \textcolor[HTML]{000000}{0.8868} & \textcolor[HTML]{000000}{4.6411} & \textcolor[HTML]{000000}{1.1396} & \textcolor[HTML]{000000}{0.8176} & 0.8189 & \textcolor[HTML]{000000}{10.9588} & \textcolor[HTML]{000000}{4.4855} & \textcolor[HTML]{0000FF}{0.7528} & \textcolor[HTML]{000000}{0.8677} & \textcolor[HTML]{000000}{83.5661} & \textcolor[HTML]{000000}{20.8422} & \textcolor[HTML]{000000}{0.7802} \\
& CPC \textcolor{gray}{\scriptsize [CVPR 2021]} & 0.8248 & \textcolor[HTML]{000000}{13.9334} & \textcolor[HTML]{000000}{4.8242} & \textcolor[HTML]{000000}{0.7324} & \textcolor[HTML]{000000}{0.8777} & \textcolor[HTML]{000000}{4.3746} & \textcolor[HTML]{000000}{1.2428} & \textcolor[HTML]{000000}{0.8079} & \textcolor[HTML]{0000FF}{0.8191} & \textcolor[HTML]{0000FF}{10.3361} & \textcolor[HTML]{000000}{4.3526} & \textcolor[HTML]{000000}{0.7502} & \textcolor[HTML]{000000}{0.8709} & \textcolor[HTML]{0000FF}{16.3723} & \textcolor[HTML]{000000}{6.2057} & \textcolor[HTML]{000000}{0.7858} \\
& MT \textcolor{gray}{\scriptsize [NeruIPS 2017]} & 0.7498 & \textcolor[HTML]{000000}{12.4297} & \textcolor[HTML]{000000}{6.9233} & \textcolor[HTML]{000000}{0.6312} & \textcolor[HTML]{000000}{0.7514} & \textcolor[HTML]{000000}{16.1934} & \textcolor[HTML]{000000}{5.7225} & \textcolor[HTML]{000000}{0.6367} & 0.8190 & \textcolor[HTML]{000000}{11.6729} & \textcolor[HTML]{0000FF}{4.2150} & \textcolor[HTML]{000000}{0.7451} & \textcolor[HTML]{000000}{0.8578} & \textcolor[HTML]{000000}{208.9158} & \textcolor[HTML]{000000}{43.4428} & \textcolor[HTML]{000000}{0.7771} \\
& UA-MT \textcolor{gray}{\scriptsize [MICCAI 2019]} & 0.7596 & \textcolor[HTML]{000000}{28.0518} & \textcolor[HTML]{000000}{9.2774} & \textcolor[HTML]{000000}{0.6442} & \textcolor[HTML]{000000}{0.7580} & \textcolor[HTML]{000000}{16.9148} & \textcolor[HTML]{000000}{5.9960} & \textcolor[HTML]{000000}{0.6453} & 0.8174 & \textcolor[HTML]{000000}{13.4336} & \textcolor[HTML]{000000}{7.4999} & \textcolor[HTML]{000000}{0.7430} & \textcolor[HTML]{000000}{0.8421} & \textcolor[HTML]{000000}{173.2721} & \textcolor[HTML]{000000}{29.3279} & \textcolor[HTML]{000000}{0.7641} \\
& Self-Training \textcolor{gray}{\scriptsize [MICCAI 2019]} & 0.7609 & \textcolor[HTML]{000000}{28.6486} & \textcolor[HTML]{000000}{9.3243} & \textcolor[HTML]{000000}{0.6435} & \textcolor[HTML]{000000}{0.8409} & \textcolor[HTML]{000000}{12.4617} & \textcolor[HTML]{000000}{4.3107} & \textcolor[HTML]{000000}{0.7557} & 0.7456 & \textcolor[HTML]{000000}{28.1040} & \textcolor[HTML]{000000}{16.7551} & \textcolor[HTML]{000000}{0.6622} & \textcolor[HTML]{000000}{0.8219} & \textcolor[HTML]{000000}{267.1231} & \textcolor[HTML]{000000}{61.3421} & \textcolor[HTML]{000000}{0.7398} \\
& MCT \textcolor{gray}{\scriptsize [MICCAI 2021]} & 0.7934 & \textcolor[HTML]{000000}{12.5166} & \textcolor[HTML]{000000}{3.5936} & \textcolor[HTML]{000000}{0.6844} & \textcolor[HTML]{000000}{0.8161} & \textcolor[HTML]{000000}{11.8576} & \textcolor[HTML]{000000}{3.5847} & \textcolor[HTML]{000000}{0.7220} & 0.7621 & \textcolor[HTML]{000000}{15.2515} & \textcolor[HTML]{000000}{5.9778} & \textcolor[HTML]{000000}{0.6735} & 0.8161 & 35.2185 & 10.5821 & 0.7220 \\
& MCT++ \textcolor{gray}{\scriptsize [MIA 2022]} & 0.7891 & \textcolor[HTML]{000000}{10.9286} & \textcolor[HTML]{000000}{2.8662} & \textcolor[HTML]{000000}{0.6712} & \textcolor[HTML]{000000}{0.8221} & \textcolor[HTML]{000000}{12.7584} & \textcolor[HTML]{000000}{4.3217} & \textcolor[HTML]{000000}{0.7256} & 0.7372 & \textcolor[HTML]{000000}{25.0444} & \textcolor[HTML]{000000}{11.2425} & \textcolor[HTML]{000000}{0.6480} & 0.8221 & 40.5077 & 12.1035 & 0.7256 \\
& DTC \textcolor{gray}{\scriptsize [AAAI 2021]} & \textcolor[HTML]{0000FF}{0.8437} & \textcolor[HTML]{000000}{9.6753} & \textcolor[HTML]{000000}{2.9989} & \textcolor[HTML]{0000FF}{0.7540} & \textcolor[HTML]{0000FF}{0.8889} & \textcolor[HTML]{0000FF}{4.2583} & \textcolor[HTML]{0000FF}{1.1094} & \textcolor[HTML]{0000FF}{0.8217} & \textcolor[HTML]{000000}{0.8025} & \textcolor[HTML]{000000}{16.7712} & \textcolor[HTML]{000000}{6.4445} & \textcolor[HTML]{000000}{0.7227} & \textcolor[HTML]{000000}{0.8336} & \textcolor[HTML]{000000}{105.2980} & \textcolor[HTML]{000000}{24.8586} & \textcolor[HTML]{000000}{0.7384} \\
& FBA \textcolor{gray}{\scriptsize [MILLN 2023]} & 0.7509 & \textcolor[HTML]{000000}{14.8192} & \textcolor[HTML]{000000}{4.7847} & \textcolor[HTML]{000000}{0.6330} & \textcolor[HTML]{000000}{0.8114} & \textcolor[HTML]{000000}{11.8519} & \textcolor[HTML]{000000}{3.8552} & \textcolor[HTML]{000000}{0.7173} & 0.7579 & \textcolor[HTML]{000000}{26.0499} & \textcolor[HTML]{000000}{11.5671} & \textcolor[HTML]{000000}{0.6695} & \textcolor[HTML]{000000}{0.8237} & \textcolor[HTML]{000000}{22.2492} & \textcolor[HTML]{000000}{9.2379} & \textcolor[HTML]{000000}{0.7263} \\
& MCF \textcolor{gray}{\scriptsize [CVPR 2023]} & 0.8363 & \textcolor[HTML]{000000}{11.4815 } & \textcolor[HTML]{000000}{3.5088} & \textcolor[HTML]{000000}{0.7435} & \textcolor[HTML]{000000}{0.8845} & \textcolor[HTML]{000000}{5.5293} & \textcolor[HTML]{000000}{1.3564} & \textcolor[HTML]{000000}{0.8129} & 0.7787 & \textcolor[HTML]{000000}{13.1007} & \textcolor[HTML]{000000}{8.2866} & \textcolor[HTML]{000000}{0.6962} & 0.8615 & 18.7532 & \textcolor[HTML]{0000FF}{5.8990} & 0.7785 \\
& BSNet \textcolor{gray}{\scriptsize [TMI 2024]} & 0.8195 & \textcolor[HTML]{000000}{10.8921 } & \textcolor[HTML]{000000}{3.3739} & \textcolor[HTML]{000000}{0.7233} & \textcolor[HTML]{000000}{0.8663} & \textcolor[HTML]{000000}{5.8229} & \textcolor[HTML]{000000}{1.7034} & \textcolor[HTML]{000000}{0.7909} & 0.7891 & \textcolor[HTML]{000000}{12.6045} & \textcolor[HTML]{000000}{5.5733} & \textcolor[HTML]{000000}{0.7218} & \textcolor[HTML]{0000FF}{0.8725} & 17.2010 & 7.1147 & 0.7895 \\
& CMF \textcolor{gray}{\scriptsize [ACMMM 2024]} & 0.8116 & \textcolor[HTML]{000000}{18.9286} & \textcolor[HTML]{000000}{5.8365} & \textcolor[HTML]{000000}{0.7070} & \textcolor[HTML]{000000}{0.8682} & \textcolor[HTML]{000000}{11.7148} & \textcolor[HTML]{000000}{3.6431} & \textcolor[HTML]{000000}{0.7889} & 0.7686 & \textcolor[HTML]{000000}{15.2001} & \textcolor[HTML]{000000}{7.3084} & \textcolor[HTML]{000000}{0.6766} & 0.8665 & 22.9040 & 7.3425 & 0.7830 \\
& PMT \textcolor{gray}{\scriptsize [ECCV 2024]} & 0.8221 & \textcolor[HTML]{000000}{14.3792 } & \textcolor[HTML]{000000}{4.3681} & \textcolor[HTML]{000000}{0.7329} & \textcolor[HTML]{000000}{0.8739} & \textcolor[HTML]{000000}{7.1231} & \textcolor[HTML]{000000}{3.9362} & \textcolor[HTML]{000000}{0.8058} & 0.8012 & \textcolor[HTML]{000000}{12.4792} & \textcolor[HTML]{000000}{4.2649} & \textcolor[HTML]{000000}{0.7215} & 0.8730 & 16.8525 & 6.9968 & \textcolor[HTML]{0000FF}{0.7890} \\
& Ours & \textcolor[HTML]{FF0000}{0.8568} & \textcolor[HTML]{FF0000}{8.7349} & \textcolor[HTML]{FF0000}{2.3675} & \textcolor[HTML]{FF0000}{0.7754} & \textcolor[HTML]{FF0000}{0.9009} & \textcolor[HTML]{FF0000}{3.7817} & \textcolor[HTML]{FF0000}{0.8105} & \textcolor[HTML]{FF0000}{0.8354} & \textcolor[HTML]{FF0000}{0.8400} & \textcolor[HTML]{FF0000}{6.9464} & \textcolor[HTML]{FF0000}{2.8431} & \textcolor[HTML]{FF0000}{0.7754} & \textcolor[HTML]{FF0000}{0.8895} & \textcolor[HTML]{FF0000}{15.4037} & \textcolor[HTML]{FF0000}{4.6630} & \textcolor[HTML]{FF0000}{0.8041} \\
\midrule
\multirow{14}{*}{20\%} & CPS \textcolor{gray}{\scriptsize [CVPR 2021]} & \textcolor[HTML]{000000}{0.8352} & \textcolor[HTML]{0000FF}{9.2916} & \textcolor[HTML]{000000}{3.5082} & \textcolor[HTML]{000000}{0.7477} & \textcolor[HTML]{0000FF}{0.8892} & \textcolor[HTML]{0000FF}{4.7745} & \textcolor[HTML]{000000}{1.3818} & \textcolor[HTML]{0000FF}{0.8222} & \textcolor[HTML]{000000}{0.8145} & \textcolor[HTML]{000000}{9.9613} & \textcolor[HTML]{000000}{2.8608} & \textcolor[HTML]{000000}{0.7482} & \textcolor[HTML]{000000}{0.8839} & \textcolor[HTML]{000000}{15.3931} & \textcolor[HTML]{000000}{6.2130} & \textcolor[HTML]{000000}{0.8206} \\
& CPC \textcolor{gray}{\scriptsize [CVPR 2021]} & \textcolor[HTML]{000000}{0.8345} & \textcolor[HTML]{000000}{10.6067} & \textcolor[HTML]{000000}{3.3828} & \textcolor[HTML]{000000}{0.7431} & \textcolor[HTML]{000000}{0.8759} & \textcolor[HTML]{000000}{5.4510} & \textcolor[HTML]{000000}{1.4505} & \textcolor[HTML]{000000}{0.8055} & \textcolor[HTML]{0000FF}{0.8358} & \textcolor[HTML]{0000FF}{7.5041} & \textcolor[HTML]{0000FF}{2.0646} & \textcolor[HTML]{0000FF}{0.7668} & \textcolor[HTML]{000000}{0.8826} & \textcolor[HTML]{000000}{15.5970} & \textcolor[HTML]{000000}{5.0921} & \textcolor[HTML]{000000}{0.8185} \\
& MT \textcolor{gray}{\scriptsize [NeruIPS 2017]} & \textcolor[HTML]{000000}{0.7723} & \textcolor[HTML]{000000}{10.3419} & \textcolor[HTML]{000000}{5.3429} & \textcolor[HTML]{000000}{0.6701} & \textcolor[HTML]{000000}{0.7827} & \textcolor[HTML]{000000}{13.5562} & \textcolor[HTML]{000000}{4.6647} & \textcolor[HTML]{000000}{0.6761} & \textcolor[HTML]{000000}{0.8174} & \textcolor[HTML]{000000}{13.4277} & \textcolor[HTML]{000000}{4.5784} & \textcolor[HTML]{000000}{0.7430} & \textcolor[HTML]{000000}{0.8577} & \textcolor[HTML]{000000}{192.9185} & \textcolor[HTML]{000000}{37.3328} & \textcolor[HTML]{000000}{0.7769} \\
& UA-MT \textcolor{gray}{\scriptsize [MICCAI 2019]} & \textcolor[HTML]{000000}{0.7879} & \textcolor[HTML]{000000}{19.3464} & \textcolor[HTML]{000000}{6.4791} & \textcolor[HTML]{000000}{0.6832} & \textcolor[HTML]{000000}{0.7809} & \textcolor[HTML]{000000}{12.2151} & \textcolor[HTML]{000000}{4.0572} & \textcolor[HTML]{000000}{0.6700} & \textcolor[HTML]{000000}{0.8231} & \textcolor[HTML]{000000}{10.7396} & \textcolor[HTML]{000000}{4.9058} & \textcolor[HTML]{000000}{0.7441} & \textcolor[HTML]{000000}{0.8492} & \textcolor[HTML]{000000}{162.4909} & \textcolor[HTML]{000000}{25.0389} & \textcolor[HTML]{000000}{0.7695} \\
& Self-Training \textcolor{gray}{\scriptsize [MICCAI 2019]} & \textcolor[HTML]{000000}{0.7986} & \textcolor[HTML]{000000}{17.1350} & \textcolor[HTML]{000000}{6.0755} & \textcolor[HTML]{000000}{0.6958} & \textcolor[HTML]{000000}{0.8579} & \textcolor[HTML]{000000}{8.2902} & \textcolor[HTML]{000000}{2.6783} & \textcolor[HTML]{000000}{0.7780} & \textcolor[HTML]{000000}{0.7659} & \textcolor[HTML]{000000}{24.3485} & \textcolor[HTML]{000000}{11.7846} & \textcolor[HTML]{000000}{0.6867} & \textcolor[HTML]{000000}{0.8293} & \textcolor[HTML]{000000}{242.9802} & \textcolor[HTML]{000000}{53.2397} & \textcolor[HTML]{000000}{0.7403} \\
& MCT \textcolor{gray}{\scriptsize [MICCAI 2021]} & \textcolor[HTML]{000000}{0.8375} & \textcolor[HTML]{000000}{9.8259} & \textcolor[HTML]{000000}{3.1555} & \textcolor[HTML]{000000}{0.7427} & \textcolor[HTML]{000000}{0.8210} & \textcolor[HTML]{000000}{12.5363} & \textcolor[HTML]{000000}{4.2293} & \textcolor[HTML]{000000}{0.7247} & \textcolor[HTML]{000000}{0.7529} & \textcolor[HTML]{000000}{15.1901} & \textcolor[HTML]{000000}{6.3527} & \textcolor[HTML]{000000}{0.6624} & 0.8210 & 11.8983 & 8.5067 & 0.7247 \\
& MCT++ \textcolor{gray}{\scriptsize [MIA 2022]} & \textcolor[HTML]{000000}{0.8314} & \textcolor[HTML]{000000}{10.2735} & \textcolor[HTML]{0000FF}{2.6074} & \textcolor[HTML]{000000}{0.7350} & \textcolor[HTML]{000000}{0.8719} & \textcolor[HTML]{000000}{5.8454} & \textcolor[HTML]{0000FF}{1.3540} & \textcolor[HTML]{000000}{0.7953} & \textcolor[HTML]{000000}{0.7699} & \textcolor[HTML]{000000}{16.5281} & \textcolor[HTML]{000000}{6.8284} & \textcolor[HTML]{000000}{0.6763} & 0.8719 & 9.5040 & 7.0112 & 0.7953 \\
& DTC \textcolor{gray}{\scriptsize [AAAI 2021]} & \textcolor[HTML]{0000FF}{0.8456} & \textcolor[HTML]{000000}{10.6364} & \textcolor[HTML]{000000}{3.4067} & \textcolor[HTML]{0000FF}{0.7613} & \textcolor[HTML]{000000}{0.8820} & \textcolor[HTML]{000000}{6.0854} & \textcolor[HTML]{000000}{1.9215} & \textcolor[HTML]{000000}{0.8109} & \textcolor[HTML]{000000}{0.8232} & \textcolor[HTML]{000000}{14.9156} & \textcolor[HTML]{000000}{5.9355} & \textcolor[HTML]{000000}{0.7529} & \textcolor[HTML]{000000}{0.8541} & \textcolor[HTML]{000000}{23.6399} & \textcolor[HTML]{000000}{14.8962} & \textcolor[HTML]{000000}{0.7712} \\
& FBA \textcolor{gray}{\scriptsize [MILLN 2023]} & \textcolor[HTML]{000000}{0.8094} & \textcolor[HTML]{000000}{16.9585} & \textcolor[HTML]{000000}{6.0262} & \textcolor[HTML]{000000}{0.7140} & \textcolor[HTML]{000000}{0.8625} & \textcolor[HTML]{000000}{7.7679} & \textcolor[HTML]{000000}{2.4202} & \textcolor[HTML]{000000}{0.7817} & \textcolor[HTML]{000000}{0.7953} & \textcolor[HTML]{000000}{17.6334} & \textcolor[HTML]{000000}{7.9428} & \textcolor[HTML]{000000}{0.7176} & \textcolor[HTML]{000000}{0.8693} & \textcolor[HTML]{000000}{13.2397} & \textcolor[HTML]{000000}{6.3297} & \textcolor[HTML]{000000}{0.7842} \\
& MCF \textcolor{gray}{\scriptsize [CVPR 2023]} & \textcolor[HTML]{000000}{0.8361} & \textcolor[HTML]{000000}{10.5889} & \textcolor[HTML]{000000}{3.3862} & \textcolor[HTML]{000000}{0.7464} & \textcolor[HTML]{000000}{0.8847} & \textcolor[HTML]{000000}{5.5127} & \textcolor[HTML]{000000}{1.4193} & \textcolor[HTML]{000000}{0.8159} & \textcolor[HTML]{000000}{0.7798} & \textcolor[HTML]{000000}{17.2018} & \textcolor[HTML]{000000}{5.0441} & \textcolor[HTML]{000000}{0.6913} & 0.8847 & 11.2019 & \textcolor[HTML]{0000FF}{3.7840} & 0.8159 \\
& BSNet \textcolor{gray}{\scriptsize [TMI 2024]} & 0.8327 & \textcolor[HTML]{000000}{10.3948 } & \textcolor[HTML]{000000}{3.5739} & \textcolor[HTML]{000000}{0.7428} & \textcolor[HTML]{000000}{0.8752} & \textcolor[HTML]{000000}{5.6190} & \textcolor[HTML]{000000}{1.5305} & \textcolor[HTML]{000000}{0.8023} & 0.8111 & \textcolor[HTML]{000000}{8.9500} & \textcolor[HTML]{000000}{2.5896} & \textcolor[HTML]{000000}{0.7354} & 0.8930 & 10.0512 & 7.1392 & 0.8220 \\
& CMF \textcolor{gray}{\scriptsize [ACMMM 2024]} & 0.8244 & \textcolor[HTML]{000000}{17.3716 } & \textcolor[HTML]{000000}{5.9603} & \textcolor[HTML]{000000}{0.7269} & \textcolor[HTML]{000000}{0.8635} & \textcolor[HTML]{000000}{2.6049} & \textcolor[HTML]{000000}{3.6752} & \textcolor[HTML]{000000}{0.7837} & 0.7986 & \textcolor[HTML]{000000}{14.3005} & \textcolor[HTML]{000000}{5.7361} & \textcolor[HTML]{000000}{0.7098} & 0.8880 & \textcolor[HTML]{0000FF}{8.9530} & 5.9581 & 0.8195 \\
& PMT \textcolor{gray}{\scriptsize [ECCV 2024]} & 0.8412 & \textcolor[HTML]{000000}{9.9738 } & \textcolor[HTML]{000000}{3.9346} & \textcolor[HTML]{000000}{0.7468} & \textcolor[HTML]{000000}{0.8802} & \textcolor[HTML]{000000}{5.8935} & \textcolor[HTML]{000000}{1.7937} & \textcolor[HTML]{000000}{0.8117} & 0.8139 & \textcolor[HTML]{000000}{11.3242} & \textcolor[HTML]{000000}{4.4902} & \textcolor[HTML]{000000}{0.7403} & \textcolor[HTML]{0000FF}{0.8945} & 9.8801 & 6.3221 & \textcolor[HTML]{0000FF}{0.8225} \\
& Ours & \textcolor[HTML]{FF0000}{0.8591} & \textcolor[HTML]{FF0000}{8.6240} & \textcolor[HTML]{FF0000}{2.9829} & \textcolor[HTML]{FF0000}{0.7788} & \textcolor[HTML]{FF0000}{0.9007} & \textcolor[HTML]{FF0000}{4.2163} & \textcolor[HTML]{FF0000}{0.9981} & \textcolor[HTML]{FF0000}{0.8334} & \textcolor[HTML]{FF0000}{0.8593} & \textcolor[HTML]{FF0000}{6.6270} & \textcolor[HTML]{FF0000}{1.6231} & \textcolor[HTML]{FF0000}{0.7945} & \textcolor[HTML]{FF0000}{0.9083} & \textcolor[HTML]{FF0000}{7.7713} & \textcolor[HTML]{FF0000}{3.6331} & \textcolor[HTML]{FF0000}{0.8384}
\\ \bottomrule
\end{tabular}
}
\label{tab:table1} 
\end{table*}

We conduct experiments on four widely used volumetric medical imaging datasets and one privately curated dataset:

\textbf{BraTS 2021 Dataset}\cite{brats2021}: This dataset provides 1,251 cases, each with preprocessed multi-parametric MRI scans (T1, T1-ce, T2, FLAIR) in a 240×240×155 isotropic (1 mm³) format. Annotations delineate three tumor sub-regions. Following the protocol in [25], we partitioned the data into 1000, 125, and 125 cases for training, validation, and testing.

\textbf{BraTS 2020 Dataset}\cite{brats2020}: This benchmark contains 369 cases with similar imaging and preprocessing standards. We divided it into training, validation, and testing sets of 295, 37, and 37 cases, respectively.

\textbf{BraTs 2023 MEN dataset}\cite{brats2023men}: This dataset, part of the BraTS 2023 challenge, focuses on meningioma segmentation. It comprises multi-institutional multiparametric MRI scans (t1w, t1c, t2w, t2f). The training set released for the challenge contains 1000 annotated cases, with annotations for meningioma sub-regions (e.g., enhancing tumor). Following a common split strategy for this dataset, we use 800 cases (80\%) for training, 100 for validation, and 100 (10\%) for testing. 

\textbf{CBCT Tooth dataset}: This dataset consists of Cone Beam Computed Tomography (CBCT) scans focused on the dental anatomy. The task is segmentation of individual teeth, which contains 260 cases. Following a typical split for this dataset, we use 8:1:1.

\textbf{IXI Dataset}\cite{IXI}: We employed the IXI dataset, a multi-site repository of brain MRI from approximately 600 healthy participants. The dataset provides T1-weighted, T2-weighted, and Proton Density (PD) images, which have been preprocessed through skull-stripping and normalization to the MNI standard space at a 1 mm³ isotropic resolution. Our evaluation focused on the segmentation performance on white and gray matter tissues within this cohort.

\subsubsection{Evaluation metrics}
To comprehensively evaluate the segmentation performance of the model, we employed metrics based on both region accuracy and boundary distance. For region-based accuracy, we utilized the Dice Similarity Coefficient and the Jaccard Index. For boundary-based distance, we used the 95th percentile Hausdorff Distance (HD) and the Average Surface Distance (ASD).

\subsubsection{Implementation details}
All experiments were conducted on NVIDIA RTX 4090 GPUs, with CUDA version 12.4 and Python version 3.9.13. The proposed BARL was implemented based on the PyTorch 1.11.0 framework. We employed the SGDW optimizer with a momentum of 0.9 and a weight decay of 5e-4 to update the model parameters. Additionally, a Cosine Annealing scheduler was utilized to adjust the learning rate. The batch size was adjusted according to the relationship between GPU memory and computational load. For the BraTS series datasets, the learning rate was gradually decreased from 0.004 to 0.00001 over 100 epochs, which included a 20-epoch warm-up period. For the IXI dataset, the learning rate was annealed from 0.002 to 0.0005 within 50 epochs. For the CBCT dataset, we set a total of 60 epochs, with the learning rate scheduled from 0.006 to 0.0001. For image enhancement, we employed a data augmentation strategy similar to the weak-to-strong approach presented in \cite{posionly}.

This study deployed a modified Attention U-Net for fundamental experiments such as ablation studies and comparisons. To accommodate the representation space alignment strategy, we incorporated a representation head in parallel with the model's segmentation head, following the same structure as \cite{posionly}: Conv $\rightarrow$ Norm $\rightarrow$ ReLU $\rightarrow$ Conv. The dimensions of the region-level and instance-level representation vectors were set to 128 \cite{centerloss}. Subsequent representation experiments will discuss in detail the effects of different representation spaces and dimensions.

For the 3D connected-component filtering in our instance-level alignment, we set the minimum volume threshold to $\tau_{\text{vol}} = 50$. We observed that the occurrence of small, spurious components substantially decreased as training stabilized. In the information maximization module, the target prior distribution $\mathbf{q}$ was computed based on the empirical class distribution of each respective dataset. Following MCT \cite{mct}, we adopt a probability softening technique to emphasize salient regions.

For a fair and direct comparison, all baseline methods were reproduced and tested under identical conditions. The reliability of our results is reinforced through a standard five-fold validation procedure for all experiments.

\subsection{Comparison With State-of-the-Art Methods}

\begin{table*}[htbp]
\centering
\caption{Ablation Study: Evaluating the Contribution of Each Module on the BraTs 2020 test set with 20\% labeled cases. Best results are bolded.}
\label{tab:ablation}
\resizebox{\textwidth}{!}{
\begin{tabular}{@{}ccc|cccc|llll@{}}
\toprule
\multicolumn{3}{c|}{\textbf{Representation Space}} & \multicolumn{4}{c|}{\textbf{Label Space}} & \multicolumn{4}{c}{\textbf{Metrics (Evaluate on BraTs 2020)}} \\
\cmidrule(lr){1-3} \cmidrule(lr){4-7} \cmidrule(lr){8-11}
\multirow{2}{*}{\textbf{region-wise}} & \multirow{2}{*}{\textbf{instance-wise}} & \multirow{2}{*}{\textbf{lesion-wise}} & \multicolumn{3}{c|}{\textbf{DPR}} & \multirow{2}{*}{\textbf{PCBC}} & \multirow{2}{*}{\textit{Dice $\uparrow$}} & \multirow{2}{*}{\textit{HD $\downarrow$}} & \multirow{2}{*}{\textit{ASD $\downarrow$}} & \multirow{2}{*}{\textit{Jaccard $\uparrow$}} \\
\cmidrule(lr){4-6}
& & & \textbf{DeepCPS} & \textbf{dirtr} & \multicolumn{1}{c|}{\textbf{IM loss}} & & & & & \\
\midrule
& & & & & & & 0.7875 & 19.3816 & 6.3948 & 0.6828 \\
\midrule
& & &\checkmark &\checkmark &\checkmark &\checkmark &0.8323 &12.8695 &5.4432 &0.7467 \\
\checkmark & & &\checkmark &\checkmark &\checkmark &\checkmark &0.8460 &10.1677 &3.9851 &0.7601 \\
& \checkmark & &\checkmark  &\checkmark  &\checkmark  &\checkmark  &0.8418 &10.6803 &3.1159 &0.7551 \\
 & & \checkmark & \checkmark &\checkmark &\checkmark &\checkmark &0.8379 &11.3298 &4.5492 &0.7492 \\
 \midrule
\checkmark & \checkmark &  & &  & &\checkmark  &0.8453 &10.7179 &3.9911 &0.7618 \\
\checkmark & \checkmark &  & \checkmark & &  & &0.8373 &11.2329 &4.6913 &0.7488 \\
\checkmark & \checkmark &  & \checkmark &\checkmark &  & &0.8452 &11.0322 &4.2489 &0.7583 \\
\checkmark & \checkmark &  & \checkmark &\checkmark &\checkmark  & &0.8531 &9.2380 &3.2342 &0.7695 \\
\midrule
\rowcolor{lightblue} 
\checkmark & \checkmark &  & \checkmark & \checkmark & \checkmark & \checkmark & \textbf{0.8591} & \textbf{8.6240} & \textbf{2.9829} & \textbf{0.7788} \\
\bottomrule
\end{tabular}
}
\end{table*}

\subsubsection{Quantitative Experiments}
To conduct a comprehensive and detailed evaluation of our proposed method, we performed extensive experiments on five distinct datasets. This evaluation benchmark comprises four publicly available, open-source datasets and one private, in-house dataset. We benchmarked our approach against thirteen recent and popular baseline methods to ensure a thorough comparison. The competing methods include several state-of-the-art semi-supervised methods: CPS \cite{cps}, MT \cite{mt}, UA-MT \cite{uamt}, Self-Training \cite{self-training}, MCT \cite{mct}, MCT++ \cite{mctv1}, DTC \cite{dtc}, FBA \cite{fba}, BSNet \cite{bsnet}, MCF \cite{mcf}, CML \cite{cml}, and PMT \cite{pmt}.

The quantitative results of this extensive comparison are detailed in Table \ref{tab:table1}. The experiments were conducted under two distinct label scarcity settings, utilizing 10\% and 20\% of the available annotated data, respectively. Across all four datasets and both label ratios, our proposed method consistently demonstrates superior performance. Specifically, under the more challenging 10\% label setting on the BraTs2020 dataset, our approach achieves a Dice score of 0.8568 and a HD of 8.7349, outperforming all other baselines. This trend of superior performance is maintained across the BraTs2021, BraTs2023 MEN, and the in-house Tooth CBCT datasets. For instance, on the Tooth CBCT data with 20\% labels, our method attains the highest Dice score of 0.9083 and the lowest HD of 7.7713.

As indicated by the red and blue highlighting in Table \ref{tab:table1}, our model consistently ranks as the top-performing or second-best method across nearly all metrics and experimental configurations. This robust and stable performance underscores the effectiveness and generalizability of our approach in handling diverse medical imaging data domains under significant label scarcity. We attribute this success to the dual consistency constraint across both feature and label spaces, which promotes the learning of robust and expressive representations by enforcing their invariance to various perturbations.

\begin{table}[htbp]
\centering
\caption{Quantitative evaluation of different baselines on IXI dataset under 5\% label ratio. Best results are bolded.}
\resizebox{0.5\textwidth}{!}{%
\begin{tabular}{
>{\columncolor[HTML]{FFFFFF}}l
>{\columncolor[HTML]{FFFFFF}}l |
>{\columncolor[HTML]{FFFFFF}}c
>{\columncolor[HTML]{FFFFFF}}c
>{\columncolor[HTML]{FFFFFF}}c
>{\columncolor[HTML]{FFFFFF}}c }
\toprule
\cellcolor[HTML]{FFFFFF} & \multicolumn{1}{c|}{\cellcolor[HTML]{FFFFFF}} & \multicolumn{4}{c}{\cellcolor[HTML]{FFFFFF}Metrics} \\ 
\multirow{-2}{*}{\cellcolor[HTML]{FFFFFF}\textbf{Ratio}} & \multicolumn{1}{c|}{\multirow{-2}{*}{\cellcolor[HTML]{FFFFFF}\textbf{Method}}} & {\color[HTML]{000000} \textit{Dice $\uparrow$}} & {\color[HTML]{000000} \textit{HD $\downarrow$}} & {\color[HTML]{000000} \textit{ASD $\downarrow$}} & \cellcolor[HTML]{FFFFFF}{\color[HTML]{000000} \textit{Jacarrd $\uparrow$}} \\ \midrule
\cellcolor[HTML]{FFFFFF} & CPS \textcolor{gray}{\scriptsize [CVPR 2021]} & {\color[HTML]{000000} 0.6963} & {\color[HTML]{000000} 3.1971} & {\color[HTML]{000000} 1.0324} & {\color[HTML]{000000} 0.5587} \\
\cellcolor[HTML]{FFFFFF} & CPC \textcolor{gray}{\scriptsize [CVPR 2021]} & {\color[HTML]{000000} 0.8729} & {\color[HTML]{000000} 1.6821} & {\color[HTML]{000000} 0.7106} & {\color[HTML]{000000} 0.7750} \\
\cellcolor[HTML]{FFFFFF} & MT \textcolor{gray}{\scriptsize [NeruIPS 2017]} & {\color[HTML]{000000} 0.7070} & {\color[HTML]{000000} 7.3586} & {\color[HTML]{000000} 1.8237} & {\color[HTML]{000000} 0.5471} \\
\cellcolor[HTML]{FFFFFF} & UA-MT \textcolor{gray}{\scriptsize [MICCAI 2019]} & {\color[HTML]{000000} 0.7292} & {\color[HTML]{000000} 6.4251} & {\color[HTML]{000000} 2.1622} & {\color[HTML]{000000} 0.5735} \\
\cellcolor[HTML]{FFFFFF} & Self-Training \textcolor{gray}{\scriptsize [MICCAI 2019]} & {\color[HTML]{000000} 0.8331} & {\color[HTML]{000000} 26.9100} & {\color[HTML]{000000} 8.0574} & {\color[HTML]{000000} 0.7257} \\
\cellcolor[HTML]{FFFFFF} & MCT \textcolor{gray}{\scriptsize [MICCAI 2021]} & {\color[HTML]{000000} 0.8356} & {\color[HTML]{000000} 2.9443} & {\color[HTML]{000000} 1.9284} & {\color[HTML]{000000} 0.7284} \\
\cellcolor[HTML]{FFFFFF} & MCT++ \textcolor{gray}{\scriptsize [MIA 2022]} & {\color[HTML]{000000} 0.8125} & {\color[HTML]{000000} 11.1332} & {\color[HTML]{000000} 4.0651} & {\color[HTML]{000000} 0.6854} \\
\cellcolor[HTML]{FFFFFF} & DTC \textcolor{gray}{\scriptsize [AAAI 2021]} & {\color[HTML]{000000} 0.8264} & {\color[HTML]{000000} 26.4904} & {\color[HTML]{000000} 6.3265} & {\color[HTML]{000000} 0.7079} \\
\cellcolor[HTML]{FFFFFF} & FBA \textcolor{gray}{\scriptsize [MILLN2023]} & {\color[HTML]{000000} 0.7899} & {\color[HTML]{000000} 7.2977} & {\color[HTML]{000000} 4.0918} & {\color[HTML]{000000} 0.6527} \\
\cellcolor[HTML]{FFFFFF} & MCF \textcolor{gray}{\scriptsize [CVPR 2023]} & {\color[HTML]{000000} 0.8456} & {\color[HTML]{000000} 3.3429} & {\color[HTML]{000000} 2.4782} & {\color[HTML]{000000} 0.7567} \\
\cellcolor[HTML]{FFFFFF} & BSNet \textcolor{gray}{\scriptsize [TMI 2024]} & {\color[HTML]{000000} 0.8678} & {\color[HTML]{000000} 2.8765} & {\color[HTML]{000000} 1.4792} & {\color[HTML]{000000} 0.7891} \\
\cellcolor[HTML]{FFFFFF} & CMF \textcolor{gray}{\scriptsize [ACMMM 2024]} & {\color[HTML]{000000} 0.8345} & {\color[HTML]{000000} 4.4682} & {\color[HTML]{000000} 2.4362} & {\color[HTML]{000000} 0.7432} \\
\cellcolor[HTML]{FFFFFF} & PMT \textcolor{gray}{\scriptsize [ECCV 2024]} & {\color[HTML]{000000} 0.8592} & {\color[HTML]{000000} 5.2497} & {\color[HTML]{000000} 3.4212} & {\color[HTML]{000000} 0.7693} \\ \cline{2-6}
\multirow{-13}{*}{\cellcolor[HTML]{FFFFFF}5 \%} & Ours & {\color[HTML]{000000} \textbf{0.8979}} & {\color[HTML]{000000} \textbf{1.0379}} & {\color[HTML]{000000} \textbf{0.6531}} & \cellcolor[HTML]{FFFFFF}{\color[HTML]{000000} \textbf{0.8082}} \\ \bottomrule
\end{tabular}%
}
\label{tab:table3}
\end{table}

The quantitative evaluation presented in Table \ref{tab:table3} highlights the superiority of our proposed method under an extreme label scarcity of 5\%. Our approach achieves state-of-the-art performance by a significant margin, delivering the best results across all four evaluation metrics. Notably, it attains a Dice score of 0.8979 and, more impressively, an exceptionally low HD of 1.0379, drastically outperforming the next-best method. This robust performance, particularly in boundary-sensitive metrics like HD and ASD, underscores the model's effectiveness and stability in low-data regimes.

\subsection{Ablation Analysis}
\subsubsection{Effects of each module}
To comprehensively evaluate the efficacy of each component within our proposed BARL framework, we conducted a detailed ablation study on the BraTS 2020 dataset, utilizing a semi-supervised setting with 20\% of the cases labeled. The results are presented in Table~\ref{tab:ablation}.

For the \textbf{Representation Space Alignment}, we observe that concurrently applying constraints at both the region-wise and instance-wise levels yields the most significant improvements. This finding not only underscores the fundamental importance of representation alignment but also highlights the necessity of enforcing consistency from diverse structural perspectives. As illustrated in Fig. \ref{fig:figure1} (b), the spatial distribution of voxels belonging to the same lesion class can be discrete and fragmented. Our experiments reveal that alignment at the fine-grained \textit{lesion-instance} level is substantially more effective than at the coarser \textit{lesion-class} level (Dice of $0.8418$ vs. $0.8379$). We postulate that this is because class-level alignment can introduce a \textit{prototype bias}, where a single prototype fails to capture the multi-component nature of the lesion, thereby leading to the learning of imbalanced representations.

In the \textbf{Label Space}, the removal of either the PCBC or the DPR module invariably leads to a degradation in performance. For the DPR module in particular, our results indicate that a synergistic combination of constraints on hard pseudo-labels and soft logits is most beneficial. Furthermore, the IM loss, which is designed to increase the confidence of the model's output and prevent it from collapsing to trivial solutions, proves to be a valuable component. Its exclusion results in a noticeable decline in segmentation accuracy (e.g., the Dice score drops from $0.8591$ to $0.8531$).

In summary, the complete BARL strategy, which integrates all proposed modules, achieves the best performance. As shown in the last row of the table, the proposed BARL obtains a Dice score of $0.8591$, a HD of $8.6240$, an ASD of $2.9829$, and a Jaccard index of $0.7788$. The fact that the removal of any single module results in a performance drop provides compelling evidence for the effectiveness and indispensability of each component within our proposed bilateral alignment framework.

\begin{table}[ht]
\centering
\caption{Performance comparison of different label space alignment tool combinations across datasets.}
\begin{tabular}{llcccc}
\toprule
Dataset & Alignment Tools & Dice$\uparrow$ & HD$\downarrow$ & ASD$\downarrow$ & Jaccard$\uparrow$ \\
\midrule
\multirow{4}{*}{BraTS2020} & MSE + CE   & 0.8568 & 8.7349 & 2.3675 & 0.7754 \\
                            & MSE + Dice & 0.8380 & 11.3126 & 3.9236 & 0.7495 \\
                            & KL + Dice  & 0.8458 & 9.6184  & 2.8874 & 0.7578 \\
                            & KL + CE    & 0.8479 & 11.8177 & 3.5542 & 0.7605 \\
\midrule
\multirow{4}{*}{BraTS2021} & MSE + CE   & 0.9009 & 3.7817 & 0.8105 & 0.8354 \\
                            & MSE + Dice & 0.8954 & 4.0228 & 0.7910 & 0.8290 \\
                            & KL + Dice  & 0.8967 & 3.5964 & 0.6918 & 0.8306 \\
                            & KL + CE    & 0.9011 & 3.3319 & 0.7944 & 0.8345 \\
\bottomrule
\end{tabular}
\label{table:tools}
\end{table}

\begin{table}[htbp]
\centering
\caption{Impact of Data Source for Representation Alignment on CBCT Tooth Segmentation Performance.}
\resizebox{0.5\textwidth}{!}{%
\begin{tabular}{cc|cccc}
\toprule
\textbf{Ratio} & \textbf{Source} & \textit{Dice $\uparrow$} & \textit{HD $\downarrow$} & \textit{ASD $\downarrow$} & \textit{Jaccard $\uparrow$} \\
\midrule
\multirow{3}{*}{20\%} & Labeled & 0.8537 & 10.5090 & 4.4692 & 0.7714 \\
 & Unlabeled & 0.8877 & 8.1379 & 5.1442 & 0.8072 \\
\rowcolor{lightblue}
 & All & 0.9083 & 7.7713 & 3.6331 & 0.8384 \\
\midrule
\multirow{3}{*}{10\%} & Labeled & 0.8777 & 18.7274 & 7.3568  & 0.8021 \\
 & Unlabeled & 0.8805 & 16.6108 & 5.4422 & 0.8039 \\
\rowcolor{lightblue}
 & All & 0.8895 & 15.4037 & 4.6630 & 0.8041 \\
\bottomrule
\end{tabular}%
}
\label{tab:source}
\end{table}

\begin{figure*}[t!]  
    \centering
    \includegraphics[width=\linewidth]{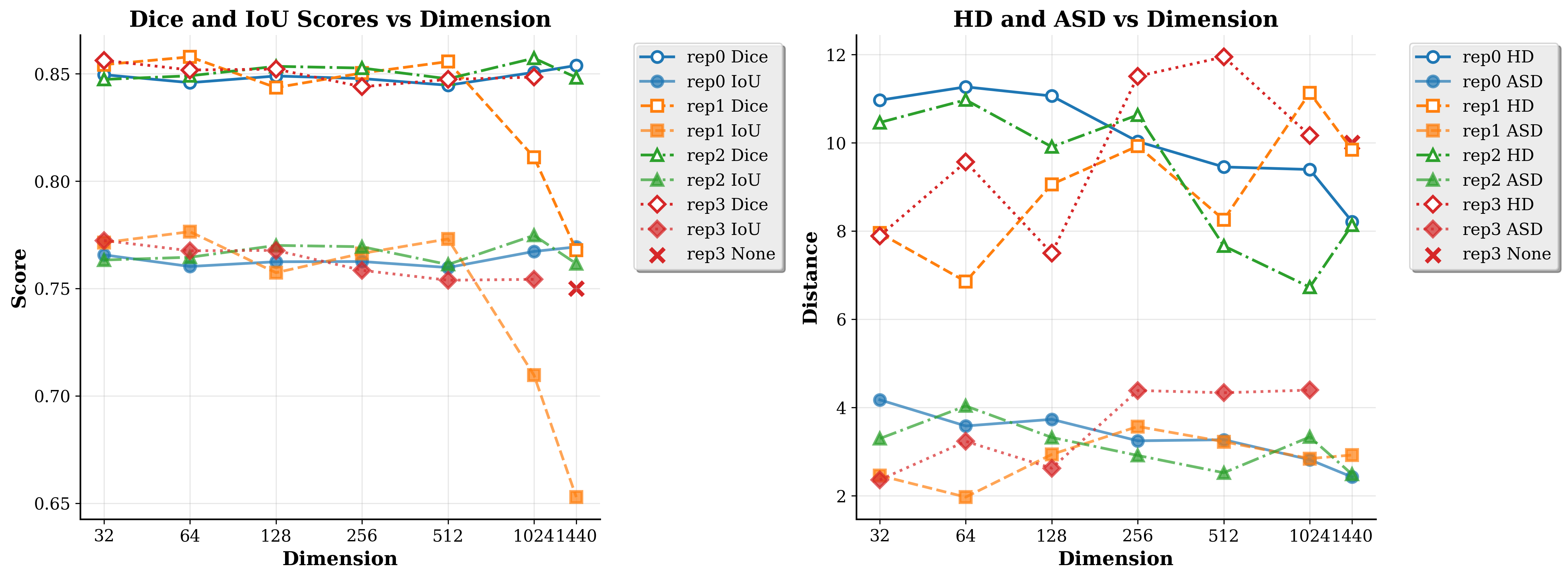}
    \captionsetup{justification=justified, singlelinecheck=false} 
    \caption{Performance under different combinations of representation spaces and dimensions on BraTs2020 in 20\% label ratio.}
    \label{fig:repdim} 
\end{figure*}
\subsubsection{Analysis of Consistency Regularization Tools}
We investigate the impact of different consistency tools for label space alignment in Table~\ref{table:tools}. Our findings suggest a trade-off between losses that operate on probability distributions and those that target spatial overlap.

We observe that consistency based on MSE, a simple and common choice \cite{mt}, does not directly optimize segmentation metrics and can result in inferior boundary quality (higher HD and ASD). In contrast, incorporating a region-based Dice loss consistently improves boundary metrics by directly optimizing for overlap. This suggests that relying on a purely distributional loss like MSE is insufficient.

Our ablation reveals that a combination of MSE and Cross-Entropy offers the best-balanced performance. The MSE term, which enforces soft distributional consistency, provides a stable and effective regularization signal. Besides, CE loss, which applies strong supervision on high-confidence pseudo-labels, a technique popularized by methods like CPS \cite{cps}. This \textit{MSE + CE} configuration achieves a remarkable result: it obtains the highest Dice/Jaccard scores on BraTS 2020 while simultaneously securing the best boundary performance on BraTS 2021. This suggests that combining a simple, soft distributional alignment (MSE) with hard pseudo-label supervision (CE) is a highly effective and robust strategy.

\subsubsection{Analysis of Representation Alignment Implementation}
We investigate a fundamental design choice in semi-supervised learning: should consistency regularization be applied to unlabeled data only, or to all data? Existing methods are divided on this topic. Many common frameworks enforce consistency exclusively on the unlabeled set to leverage its scale \cite{self, uamt}. In contrast, some works suggest the potential benefits of using the entire dataset \cite{posionly}. To address this question, we conduct a controlled experiment on the CBCT Tooth Segmentation dataset, with the results presented in Table~\ref{tab:source}.

The results demonstrate that applying representation alignment to \textbf{all} data (both labeled and unlabeled) yields the most significant performance gains across both supervision ratios. For instance, under the 20\% labeled data regime, enforcing consistency on All data achieves a Dice score of 0.9083. This represents a substantial improvement over applying it to Unlabeled data alone (0.8877) and far surpasses the performance when applied only to Labeled data (0.8537). This trend is consistently observed in the more challenging 10\% labeled data scenario, where the All data strategy again achieves the highest scores in all metrics, reinforcing the robustness of this conclusion.

We find that including labeled data in the consistency loss is surprisingly effective. Our hypothesis is that the labeled and unlabeled data play complementary roles. The large set of unlabeled data enables the model to learn a robust and consistent representation. The small set of labeled data, in turn, provides stable semantic anchors. Enforcing consistency on these anchors prevents the model from drifting due to noisy signals from the unlabeled set. This anchoring effect grounds the feature learning process to the ground-truth semantics, stabilizing the training and resulting in a more accurate model.

\subsubsection{Analysis of Representation Space and Dimension}
We conduct an analytical study to investigate the impact of the representation space and its dimensionality on the efficacy of our alignment strategy. Using an AttentionUNet \cite{attentionunet} backbone, we apply the representation alignment at four distinct architectural locations (\textit{rep0-3}) with feature dimensions ranging from 32 to 1440. The results are presented in Fig.~\ref{fig:repdim}.

\paragraph{Impact of Dimensionality}
Figure~\ref{fig:repdim} (left) shows a clear trend: segmentation performance, measured by Dice and IoU, degrades as the representation dimension increases. This effect becomes prominent for dimensions larger than 512. We attribute this to the curse of dimensionality \cite{curse}, where optimizing features in an excessively high-dimensional space can lead to overfitting with limited data. The trend is most evident in the \textit{rep3} space, where increasing the dimension to 1440 results in a complete training failure. This investigation suggests that a more compact and efficient representation is preferable to a high-dimensional, potentially sparse one for this task.

\paragraph{Analysis of Representation Space Efficacy}
We find that the choice of representation space for alignment is critical. Our experiments consistently show the best performance is achieved when applying the alignment loss at intermediate layers: specifically, the 3rd encoder layer (\textit{rep0}) and 3rd decoder layer (\textit{rep2}). We hypothesize this is because these layers offer a good trade-off between high-level semantic features and sufficient spatial detail, making them ideal for our alignment objective.

Conversely, experimental results reveal that both the bottleneck (\textit{rep1}) and the final decoder layer (\textit{rep3}) are poor choices for representation alignment. The bottleneck representation is too coarse and lacks the spatial fidelity required for precise boundaries, which is reflected in its poorer HD and ASD scores (Fig.~\ref{fig:repdim}, right). On the other hand, features from the final decoder layer (\textit{rep3}) are already highly specialized for the final pixel-wise prediction. We observe that imposing an additional alignment constraint at this stage interferes with the main segmentation task, leading to a significant drop in performance.

\paragraph{Conclusion}
In summary, our findings yield two key insights. First, an intermediate representation dimension, optimally in the range of 128 to 512, provides the most effective balance between expressiveness and optimizability. Second, the most effective location for applying representation alignment is within the intermediate encoder or decoder layers, where features are both semantically rich and spatially aware. Interestingly, this empirical finding aligns with the theoretical derivation presented in~\cite{information}. Additinaly, the bottleneck and final-layer features are less suitable due to a lack of spatial detail or over-specialization, respectively. Therefore, the success of representation alignment hinges on a judicious co-design of both the feature space and its dimensionality.

\begin{table}[htbp]
\centering
\caption{Different backbone comparison within  the proposed BARL framweork in the condition of BraTs 2023 20\% label ratio.}
\resizebox{0.5\textwidth}{!}{ 
\begin{tabular}{l| c| *4{c}| c |c}
\toprule
\multirow{2}{*}{\textbf{Backbone}} & \multirow{2}{*}{\textbf{Type}} & \multicolumn{4}{c|}{\textbf{Metrics}} & \textbf{FLOPs} &\textbf{Params}\\
                          &                       & Dice$\uparrow$   & HD$\downarrow$      & ASD$\downarrow$     & Jacarrd$\uparrow$ &(G)         & (M)              \\
\midrule
Swin-T           & Trans.                & 0.8342 & 8.8202  & 2.2781  & 0.7619  &  54.96     &      32.26           \\
SegViT                    & Trans.                & 0.7319 & 13.9254 & 4.4337  & 0.6122  &      82.43          &      63.94   \\\midrule
U-Net                     & CNN                   & 0.8574 & 5.3715  & 1.0639  & 0.7937  &      128.24        &     6.83        \\
Vnet                      & CNN                   & 0.5833 & 30.7135 & 16.7235 & 0.4730  &    68.49        &    9.44       \\
DeepLabV3                 & CNN                   & 0.7982 & 10.6450 & 3.1473  & 0.7106  &     984.49         &     68.75   \\
UperNet                   & CNN                   & 0.8173 & 8.8207  & 2.5980  & 0.7436  &    184.46         &    21.76       \\
AttentionUnet             & CNN                   & 0.8593 & 6.6270  & 1.6231  & 0.7945  &    241.09        &   15.40      \\
\rowcolor{lightblue} ResUnet                   & CNN                   & 0.8786 & 5.1654  & 0.8827  & 0.8181  &    189.56      &    12.84          \\
\bottomrule
\end{tabular}
}
\label{tab:backbone}
\end{table}

\begin{table}[htbp]
\centering
\caption{Different architectures comparison within the proposed BARL algorithm in the condition of BraTs 2023 10\% label ratio. $CCT_{noise}$ and $CCT_{dropout}$ refer to \cite{cct} with two different strategies. MCT refers to \cite{mct} framework.}
\resizebox{0.5\textwidth}{!}{ 
\begin{tabular}{l| *4{c}| c}
\toprule
\multirow{2}{*}{\textbf{Architecture}} & \multicolumn{4}{c|}{\textbf{Metrics}} & \textbf{Training Speed} \\
                                 & Dice$\uparrow$   & HD$\downarrow$   & ASD$\downarrow$   & Jacarrd$\uparrow$ & (mins/epoch)  \\
\midrule
Mean Teacher                       & 0.8204 & 7.7010  & 3.1516  & 0.7463  &  18.25    \\
Mean Teacher($Uncertainty$)                           & 0.8275 & 12.3428 & 4.9732  & 0.7497  &  19.97    \\
FixMatch                            & 0.7887 & 10.7301  & 3.3393  & 0.6952  &  18.12  \\
Singleton($MCT$)                             & 0.8318 & 10.0130 & 3.2604 & 0.7612  &  15.23    \\
Singleton($CCT_{noise}$)                      & 0.8283 & 8.3960 & 2.5213  & 0.7541  &  16.42   \\
Singleton($CCT_{dropout}$)                          & 0.8423 & 7.8895  & 1.9798  & 0.7729  &  16.63  \\
Co-Training(symmetric)                   & 0.7973 & 11.9982  & 4.0864 & 0.7066  &  17.32   \\
\rowcolor{lightblue} Co-Training(asymmetric)     & 0.8400 & 6.9464  & 2.8431  & 0.7754  &  17.98   \\
\bottomrule
\end{tabular}}
\label{tab:architecture}
\end{table}
\subsection{In-depth Evaluation}
\subsubsection{Effects of different backbones}
The performance of medical image segmentation is heavily dependent on the choice of the backbone architecture, especially under conditions of label scarcity. We study the impact of the backbone architecture by integrating our BARL framework with various CNN and Transformer models (Table~\ref{tab:backbone}). In the low-data regime of 20\% labels, we find that CNN-based backbones consistently outperform their Transformer-based counterparts. This is likely due to the strong inductive biases of convolutions, which are highly advantageous when labeled data is scarce.

Among the CNNs, the U-Net architectural family is the most effective. \textbf{ResUnet} achieves the best performance across all metrics (Dice 0.8786, HD 5.1654), and other variants like \textbf{AttentionUnet} also yield competitive results. We note that this performance is not simply a function of model size; parameter-heavy models like DeepLabV3 and SegViT deliver suboptimal results for their computational cost. The standard U-Net, for instance, is highly efficient, achieving strong performance with only 6.83M parameters. These results suggest that the U-Net paradigm provides an optimal trade-off between accuracy and efficiency for our semi-supervised framework.

\subsubsection{Comparative Analysis of Semi-Supervised Architectures}
To comprehensively evaluate the robustness and compatibility of our proposed BARL algorithm, we integrated it into four mainstream semi-supervised learning frameworks \cite{cps, fixmatch, mctv1, self}, as illustrated in Fig.~3. This analysis aims to connect the theoretical advantages and disadvantages of each architectural style with their empirical performance, with quantitative results detailed in Table \ref{tab:architecture}.

\paragraph{Mean Teacher (MT) and FixMatch}
The \textbf{Mean Teacher (MT) style (a)}, which leverages an Exponential Moving Average teacher to provide stable pseudo-labels \cite{mt, interteach, mtans}, establishes a solid performance baseline. As shown in Table~VII, it achieves a Dice score of 0.8204 with a training time of 18.25 minutes per epoch. However, its accuracy is surpassed by more advanced architectures. The \textbf{FixMatch style (b)}, relying on consistency between weak and strong augmentations \cite{fixmatch, mixmatch}, yielded the weakest results in our experiments, with the lowest Dice (0.7887) and Jaccard (0.6952) scores, despite being the fastest to train (18.12 mins/epoch). This outcome aligns with the known limitation of FixMatch: its high sensitivity to augmentation strategy and thresholding can lead to confirmation bias, hindering performance on complex medical imaging tasks.

\paragraph{Singleton Architectures}
The \textbf{Singleton style (d)} employs variations of a single model structure (one or multiple encoder/decoder), simplifying the training pipeline \cite{mct, mctv1, crossmatch, cct}. While basic singleton models can be prone to overfitting, our results demonstrate that advanced variants are highly competitive. Notably, the \textbf{Singleton(CCT\_dropout)} variant \cite{cct} delivered exceptional performance, achieving the highest Dice score and the best ASD among all tested architectures. Furthermore, this style proved to be computationally efficient, with training speeds ranging from 15.23 to 16.63 mins/epoch, faster than most other SSIS frameworks.

\paragraph{Co-Training Architectures}
The \textbf{Co-Training style (c)} utilizes disagreement between two distinct models to improve generalization and reduce error accumulation \cite{cps, cct2}. Our experiments validate its theoretical strengths. As presented in Table \ref{tab:architecture}, the \textbf{Co-Training(asymmetric)} configuration, which employs an asymmetric weak-to-strong augmentation strategy, achieved top-tier results, presenting the best HD and Jaccard scores, along with the second-highest Dice score. Contrastively, \textbf{Co-Training(symmetric)} applying symmetric strong augmentations to both branches led to a dramatic performance collapse. This confirms that creating sufficient view-disagreement through appropriate weak-to-strong data augmentation is essential to the Co-Training paradigm.

\paragraph{Conclusion}
In summary, our comparative analysis reveals a nuanced trade-off between different semi-supervised architectures when integrated with the BARL algorithm. While Mean Teacher offers a reasonable baseline, the state-of-the-art performance is led by two frameworks. The \textbf{Singleton(CCT\_dropout)} architecture excels in volumetric overlap and surface distance metrics (best Dice and ASD) while being computationally efficient. Concurrently, the \textbf{Co-Training(w aug)} framework demonstrates superior performance in boundary delineation and overall segmentation similarity (best HD and Jaccard). Given its slightly superior performance on the primary Dice metric and faster training speed, the Singleton(CCT\_dropout) presents a compelling choice. However, for applications where boundary accuracy is paramount, the Co-Training framework remains the optimal selection.

\begin{figure}[h]  
    \centering
    \includegraphics[width=0.9\linewidth]{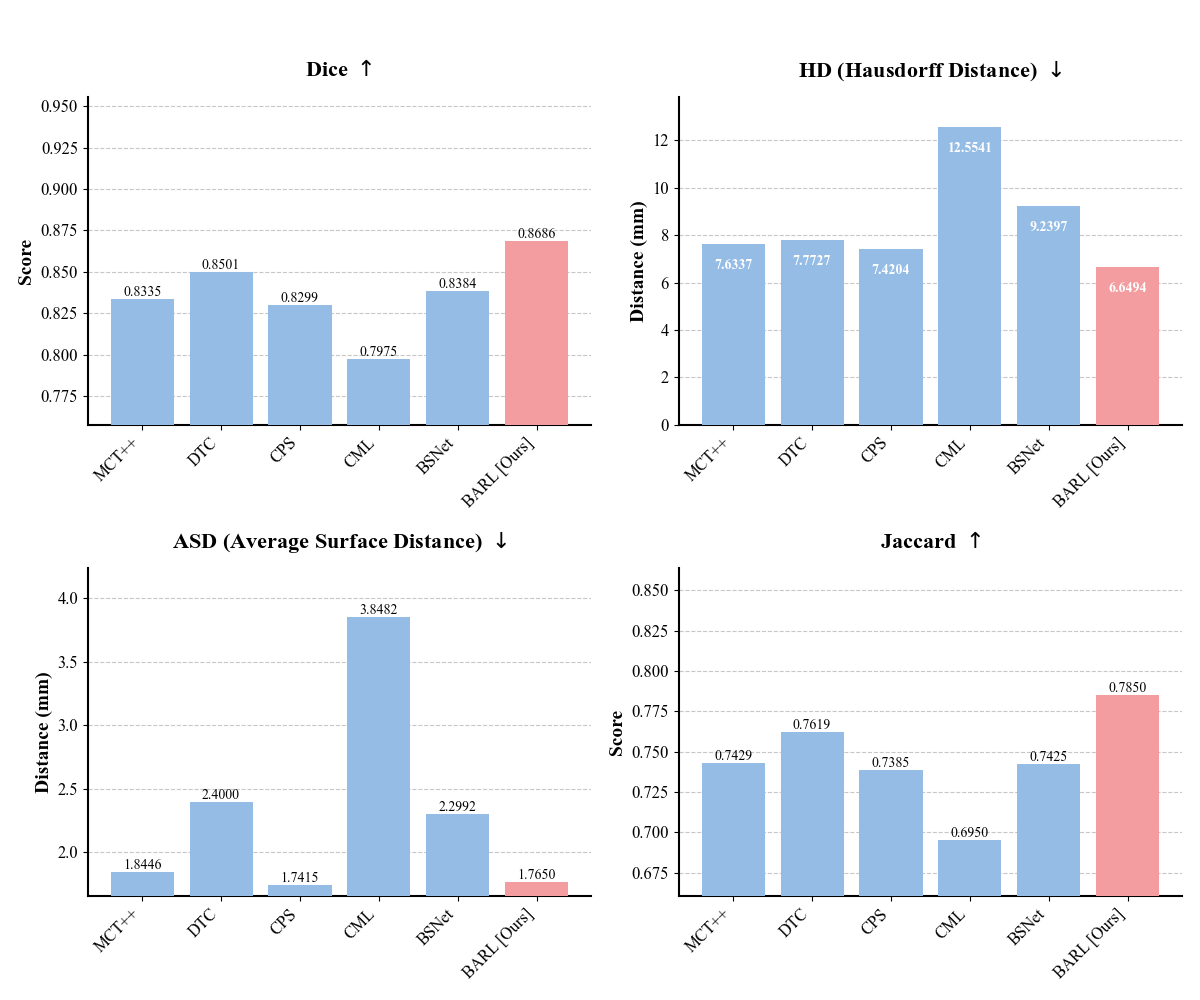}
    \captionsetup{justification=justified} 
    \caption{External data validation on BraTs2021 dataset using well-trained model on BraTs2020 dataset within 10\% ratio.}
    \label{fig:figure5} 
\end{figure}

\subsubsection{External Dataset Validation}
SSMIS typically suffers from poor generalization \cite{mct}, a challenge rooted in the intricate nature of pathological features and the severe scarcity of annotated data. To validate the generalization capability of our proposed method, we conducted a cross-dataset evaluation. The model was first trained on the BraTs2020 dataset, utilizing only a 10\% ratio of labeled data, and subsequently evaluated on the BraTs2021 test set, which serves as an out-of-distribution superset. For a fair comparison, we benchmarked BARL against five state-of-the-art semi-supervised methods: MCT++ \cite{mctv1}, DTC \cite{dtc}, CPS \cite{cps}, CML \cite{cml}, and BSNet \cite{bsnet}.

The quantitative results, presented in Fig. \ref{fig:figure5}, validate the superior generalization capability of our proposed method, BARL. When evaluated on the external dataset, BARL consistently outperforms all five competing methods across every metric. Its generalization advantage can be attributed to its dual-space regularization design, which jointly constrains the output predictions and the latent feature distributions. This sustained high performance indicates that BARL learns more robust and transferable features from limited annotations, affirming its strong potential for reliable deployment in diverse, multi-institutional clinical settings.

\section{Discussion and Conclusion}




In this paper, we reformulated semi-supervised volumetric segmentation as a dual-space constraint problem and introduced \textbf{BARL}, a novel framework that enforces synergistic constraints in both the representation space and the label space. Extensive experiments demonstrate that BARL establishes a new state-of-the-art on diverse 3D medical imaging datasets, spanning meningioma, glioma, CBCT teeth, and brain structures, under various levels of supervision. Our method surpasses 13 classic baselines, highlighting its effectiveness and robustness across different medical domains. Furthermore, successful validation on an external dataset substantiates the strong generalization capability of our method.

Subsequently, detailed ablation studies are conducted to meticulously validate the efficacy of each constituent module.
Within the representation space, our ablations elucidate the synergistic effect of aligning features at both the regional and lesion-instance levels, thereby substantiating the necessity of a coarse-to-fine constraint strategy.
Notably, we investigate the comparative efficacy of lesion-category versus lesion-instance alignment, with empirical results indicating that instance-level alignment yields superior performance gains and highlights the adverse impact of prototype shift at the category level.
Moreover, experiments concerning the dimensionality of the representation space address the curse of dimensionality and affirm the significance of feature richness \& expressiveness. Within the label space, the \textbf{DPR} module, motivated by a co-training architecture and multi-level decoders, regularizes the label distribution in conjunction with the overarching algorithmic framework and backbone network.
Concurrently, the \textbf{PCBC} module addresses the cognitive dissonance between the dual branches by rectifying the model's perception through the correction of soft uncertainty in areas of disagreement.

Furthermore, we present a series of extended experiments to enrich the theoretical validation and experimental foundation of the semi-supervised image segmentation field.
A comparative analysis of different backbones reveals that the UNet architecture remains highly effective for medical image segmentation tasks.
We also integrated the BARL strategy into various SSIS frameworks, discovering that a co-training architecture, when coupled with appropriate data augmentation, delivers optimal results.

In conclusion, BARL effectively addresses the critical challenge of label scarcity in 3D medical image segmentation, establishing a powerful new baseline and contributing rich experimental insights to the community. In the future, we will explore more advanced alignment strategies and integrate complementary techniques such as domain adaptation and contrastive learning to further advance the frontiers of semi-supervised medical image segmentation.


\begin{thebibliography}{00}\leftskip1pc

\bibitem{intro1} D. L. Pham, C. Xu, and J. L. Prince, "Current methods in medical image segmentation," \emph{Annual review of biomedical engineering}, vol. 2, no. 1, pp. 315--337, 2000.


\bibitem{intro2} R. Wang, T. Lei, R. Cui, B. Zhang, H. Meng, and A. K. Nandi, "Medical image segmentation using deep learning: A survey," \emph{IET image processing}, vol. 16, no. 5, pp. 1243--1267, 2022.


\bibitem{intro3} V. Maik, M. Naheem, K. Ram, M. Lakshmanan, M. Sivaprakasam, et al., "A Hybrid-Layered System for Image-Guided Navigation and Robot Assisted Spine Surgery," \emph{arXiv preprint arXiv}:2406.04644
        
        
        
        
        
        
        
        , 2024.

\bibitem{intro4} R. Azad, E. K. Aghdam, A. Rauland, Y. Jia, A. H. Avval, A. Bozorgpour, S. Karimijafarbigloo, J. P. Cohen, E. Adeli, and D. Merhof, "Medical image segmentation review: The success of u-net," \emph{IEEE Transactions on Pattern Analysis and Machine Intelligence}, 2024.

\bibitem{intro_ultra} S. Liu, Y. Wang, X. Yang, B. Lei, L. Liu, S. X. Li, D. Ni, and T. Wang, "Deep learning in medical ultrasound analysis: a review," \emph{Engineering}, vol. 5, no. 2, pp. 261--275, 2019.

\bibitem{intro_mri} F. Knoll, K. Hammernik, C. Zhang, S. Moeller, T. Pock, D. K. Sodickson, and M. Akcakaya, "Deep-learning methods for parallel magnetic resonance imaging reconstruction: A survey of the current approaches, trends, and issues," \emph{IEEE signal processing magazine}, vol. 37, no. 1, pp. 128--140, 2020.


\bibitem{intro_ct} M. M. Lell and M. Kachelrie{\ss}, "Recent and upcoming technological developments in computed tomography: high speed, low dose, deep learning, multienergy," \emph{Investigative radiology}, vol. 55, no. 1, pp. 8--19, 2020.

\bibitem{intro_lesion} J. Cho, K.-S. Park, M. Karki, E. Lee, S. Ko, J. K. Kim, D. Lee, J. Choe, J. Son, M. Kim, et al., "Improving sensitivity on identification and delineation of intracranial hemorrhage lesion using cascaded deep learning models," \emph{Journal of digital imaging}, vol. 32, pp. 450--461, 2019.


\bibitem{intro_plan} G. Samarasinghe, M. Jameson, S. Vinod, M. Field, J. Dowling, A. Sowmya, and L. Holloway, "Deep learning for segmentation in radiation therapy planning: a review," \emph{Journal of Medical Imaging and Radiation Oncology}, vol. 65, no. 5, pp. 578--595, 2021.

\bibitem{intro_anno} S. Wang, C. Li, R. Wang, Z. Liu, M. Wang, H. Tan, Y. Wu, X. Liu, H. Sun, R. Yang, et al., "Annotation-efficient deep learning for automatic medical image segmentation," \emph{Nature communications}, vol. 12, no. 1, pp. 5915, 2021.

\bibitem{effective} Q. Qiao, M. Qu, W. Wang, B. Jiang, and Q. Guo, "Effective Global Context Integration for Lightweight 3D Medical Image Segmentation," \emph{IEEE Transactions on Circuits and Systems for Video Technology}, 2024.

\bibitem{interteach} Z. Zhang, Q. Ma, Y. Zhang, Z. Chen, J. Chen, and H. Zheng, "InterTeach: A Novel Approach for Semi-Supervised Medical Image Segmentation Using Cooperative Teacher-Student Networks," \emph{IEEE Transactions on Circuits and Systems for Video Technology}, 2024.


\bibitem{intro_bias} Q. Yang, Z. Chen, and Y. Yuan, "Hierarchical bias mitigation for semi-supervised medical image classification," \emph{IEEE Transactions on Medical Imaging}, vol. 42, no. 8, pp. 2200--2210, 2023.


\bibitem{intro5} Y. Liu, Y. Tian, C. Wang, Y. Chen, F. Liu, V. Belagiannis, and G. Carneiro, "Translation consistent semi-supervised segmentation for 3d medical images," \emph{IEEE Transactions on Medical Imaging}, 2024.


\bibitem{cct} Y. Ouali, C. Hudelot, and M. Tami, "Semi-supervised semantic segmentation with cross-consistency training," \emph{Proceedings of the IEEE/CVF conference on computer vision and pattern recognition}, pp. 12674--12684, 2020.


\bibitem{dtc} X. Luo, J. Chen, T. Song, and G. Wang, "Semi-supervised medical image segmentation through dual-task consistency," \emph{Proceedings of the AAAI conference on artificial intelligence}, vol. 35, no. 10, pp. 8801--8809, 2021.

\bibitem{cct2} F. Zhang, H. Liu, J. Wang, J. Lyu, Q. Cai, H. Li, J. Dong, and D. Zhang, "Cross co-teaching for semi-supervised medical image segmentation," \emph{Pattern Recognition}, vol. 152, pp. 110426, 2024.

\bibitem{cps} X. Chen, Y. Yuan, G. Zeng, and J. Wang, "Semi-supervised semantic segmentation with cross pseudo supervision," \emph{Proceedings of the IEEE/CVF conference on computer vision and pattern recognition}, pp. 2613--2622, 2021.

\bibitem{mtans} G. Chen, J. Ru, Y. Zhou, I. Rekik, Z. Pan, X. Liu, Y. Lin, B. Lu, and J. Shi, "MTANS: multi-scale mean teacher combined adversarial network with shape-aware embedding for semi-supervised brain lesion segmentation," \emph{NeuroImage}, vol. 244, pp. 118568, 2021.


\bibitem{consist} Y. Fan, A. Kukleva, D. Dai, and B. Schiele, "Revisiting consistency regularization for semi-supervised learning," \emph{International Journal of Computer Vision}, vol. 131, no. 3, pp. 626--643, 2023.


\bibitem{smooth} Y. Wu, Z. Wu, Q. Wu, Z. Ge, and J. Cai, "Exploring smoothness and class-separation for semi-supervised medical image segmentation," \emph{International conference on medical image computing and computer-assisted intervention}, pp. 34--43, 2022.


\bibitem{fixmatch} K. Sohn, D. Berthelot, N. Carlini, Z. Zhang, H. Zhang, C. A. Raffel, E. D. Cubuk, A. Kurakin, and C.-L. Li, "Fixmatch: Simplifying semi-supervised learning with consistency and confidence," \emph{Advances in neural information processing systems}, vol. 33, pp. 596--608, 2020.

\bibitem{remixmatch} D. Berthelot, N. Carlini, E. D. Cubuk, A. Kurakin, K. Sohn, H. Zhang, and C. Raffel, "Remixmatch: Semi-supervised learning with distribution alignment and augmentation anchoring," \emph{arXiv preprint arXiv}:1911.09785
        
        
        
        , 2019.

\bibitem{mixmatch} D. Berthelot, N. Carlini, I. Goodfellow, N. Papernot, A. Oliver, and C. A. Raffel, "Mixmatch: A holistic approach to semi-supervised learning," \emph{Advances in neural information processing systems}, vol. 32, 2019.


\bibitem{sun2024semi} B. Sun, K. Li, J. Liu, Z. Sun, X. Wang, H. Xue, A. Hao, S. Li, and Y. Xiao, "Semi-Supervised Medical Image Segmentation with Cross-View Consistency and Contrastive Learning," \emph{2024 IEEE International Conference on Bioinformatics and Biomedicine (BIBM)}, pp. 2446--2453, 2024.



\bibitem{uda} Q. Xie, Z. Dai, E. Hovy, T. Luong, and Q. Le, "Unsupervised data augmentation for consistency training," \emph{Advances in neural information processing systems}, vol. 33, pp. 6256--6268, 2020.

\bibitem{ijcnn} E. Arazo, D. Ortego, P. Albert, N. E. O’Connor, and K. McGuinness, "Pseudo-labeling and confirmation bias in deep semi-supervised learning," \emph{2020 International joint conference on neural networks (IJCNN)}, pp. 1--8, 2020.


\bibitem{cross} N. Zhang, F. Xiao, J. Hou, R. Zhao, X. Zhang, and R. Feng, "Cross-Image Distillation for Semi-Supervised Semantic Segmentation," \emph{ICASSP 2024-2024 IEEE International Conference on Acoustics, Speech and Signal Processing (ICASSP)}, pp. 6745--6749, 2024.



\bibitem{posionly} I. Alonso, A. Sabater, D. Ferstl, L. Montesano, and A. C. Murillo, "Semi-supervised semantic segmentation with pixel-level contrastive learning from a class-wise memory bank," \emph{Proceedings of the IEEE/CVF international conference on computer vision}, pp. 8219--8228, 2021.

\bibitem{pseudoseg} Y. Zou, Z. Zhang, H. Zhang, C.-L. Li, X. Bian, J.-B. Huang, and T. Pfister, "Pseudoseg: Designing pseudo labels for semantic segmentation," \emph{arXiv preprint arXiv}:2010.09713, 2020.

\bibitem{centerloss} Y. Wu, X. Li, and Y. Zhou, "Uncertainty-aware representation calibration for semi-supervised medical imaging segmentation," \emph{Neurocomputing}, vol. 595, pp. 127912, 2024.


\bibitem{bsnet} A. He, T. Li, J. Yan, K. Wang, and H. Fu, "Bilateral supervision network for semi-supervised medical image segmentation," \emph{IEEE Transactions on Medical Imaging}, vol. 43, no. 5, pp. 1715--1726, 2023.

\bibitem{cml} S. Wu, X. Wei, X. Chen, Y. Ren, J. He, and X. Pu, "Cross-View Mutual Learning for Semi-Supervised Medical Image Segmentation," \emph{Proceedings of the 32nd ACM International Conference on Multimedia}, pp. 9253--9261, 2024.


\bibitem{pmt} N. Gao, S. Zhou, L. Wang, and N. Zheng, "PMT: Progressive Mean Teacher via Exploring Temporal Consistency for Semi-Supervised Medical Image Segmentation," \emph{European Conference on Computer Vision}, pp. 144--160, 2024.


\bibitem{contras1}
H. Basak and Z. Yin, “Pseudo-label guided contrastive learning for semi-supervised medical image segmentation,” in \emph{Proc. IEEE/CVF Conf. Comput. Vis. Pattern Recognit.}, pp. 19786–19797, 2023.

\bibitem{contras2}
Y. Wang, H. Wang, Y. Shen, J. Fei, W. Li, G. Jin, L. Wu, R. Zhao, and X. Le, “Semi-supervised semantic segmentation using unreliable pseudo-labels,” in \emph{Proc. IEEE/CVF Conf. Comput. Vis. Pattern Recognit.}, pp. 4248–4257, 2022.


\bibitem{self-training} W. Bai, O. Oktay, M. Sinclair, H. Suzuki, M. Rajchl, G. Tarroni, B. Glocker, A. King, P. M. Matthews, and D. Rueckert, "Semi-supervised learning for network-based cardiac MR image segmentation," \emph{Medical Image Computing and Computer-Assisted Intervention- MICCAI 2017: 20th International Conference, Quebec City, QC, Canada, September 11-13, 2017, Proceedings, Part II 20}, pp. 253--260, 2017.


\bibitem{labelbias}
E. Arazo, D. Ortego, P. Albert, N. E. O’Connor, and K. McGuinness, “Pseudo-labeling and confirmation bias in deep semi-supervised learning,” in \emph{2020 International Joint Conference on Neural Networks (IJCNN)}, 2020, pp. 1–8. IEEE.



\bibitem{mt}
A. Tarvainen and H. Valpola, “Mean teachers are better role models: Weight-averaged consistency targets improve semi-supervised deep learning results,” \emph{Advances in Neural Information Processing Systems}, vol. 30, 2017.



\bibitem{segvit} B. Zhang, Z. Tian, Q. Tang, X. Chu, X. Wei, C. Shen, et al., "Segvit: Semantic segmentation with plain vision transformers," \emph{Advances in Neural Information Processing Systems}, vol. 35, pp. 4971--4982, 2022.


\bibitem{swin} Z. Liu, Y. Lin, Y. Cao, H. Hu, Y. Wei, Z. Zhang, S. Lin, and B. Guo, "Swin transformer: Hierarchical vision transformer using shifted windows," \emph{Proceedings of the IEEE/CVF international conference on computer vision}, pp. 10012--10022, 2021.


\bibitem{attentionunet} O. Oktay, J. Schlemper, L. Le Folgoc, M. Lee, M. Heinrich, K. Misawa, K. Mori, S. McDonagh, N. Y. Hammerla, B. Kainz, et al., ``Attention U-Net: Learning where to look for the pancreas,'' \emph{arXiv preprint arXiv}:1804.03999
        
        
        
        
        
        
        
        , 2018.

        


\bibitem{vnet} F. Milletari, N. Navab, and S.-A. Ahmadi, "V-net: Fully convolutional neural networks for volumetric medical image segmentation," \emph{2016 fourth international conference on 3D vision (3DV)}, pp. 565--571, 2016.

\bibitem{cnntrans} L. Wu, M. Zhang, Y. Piao, Z. Yao, W. Sun, F. Tian, and H. Lu, "CNN-Transformer Rectified Collaborative Learning for Medical Image Segmentation," \emph{IEEE Transactions on Circuits and Systems for Video Technology}, 2024.



\bibitem{unet} O. Ronneberger, P. Fischer, and T. Brox, "U-net: Convolutional networks for biomedical image segmentation," \emph{Medical image computing and computer-assisted intervention--MICCAI 2015: 18th international conference, Munich, Germany, October 5-9, 2015, proceedings, part III 18}, pp. 234--241, 2015.



\bibitem{deeplab} L.-C. Chen, G. Papandreou, F. Schroff, and H. Adam, "Rethinking atrous convolution for semantic image segmentation," \emph{arXiv preprint arXiv}:1706.05587, 2017.

\bibitem{upernet} T. Xiao, Y. Liu, B. Zhou, Y. Jiang, and J. Sun, "Unified perceptual parsing for scene understanding," \emph{Proceedings of the European conference on computer vision (ECCV)}, pp. 418--434, 2018.

\bibitem{resunet} F. I. Diakogiannis, F. Waldner, P. Caccetta, and C. Wu, "ResUNet-a: A deep learning framework for semantic segmentation of remotely sensed data," \emph{ISPRS Journal of Photogrammetry and Remote Sensing}, vol. 162, pp. 94--114, 2020.


\bibitem{curse} I. V. Oseledets and E. E. Tyrtyshnikov, "Breaking the curse of dimensionality, or how to use SVD in many dimensions," \emph{SIAM Journal on Scientific Computing}, vol. 31, no. 5, pp. 3744--3759, 2009.
        
\bibitem{pseudoalign} J. Hu, C. Chen, L. Cao, S. Zhang, A. Shu, G. Jiang, and R. Ji, "Pseudo-label alignment for semi-supervised instance segmentation," \emph{Proceedings of the IEEE/CVF International Conference on Computer Vision}, pp. 16337--16347, 2023.


\bibitem{auxiliary} D. Kwon and S. Kwak, "Semi-supervised semantic segmentation with error localization network," \emph{Proceedings of the IEEE/CVF conference on computer vision and pattern recognition}, pp. 9957--9967, 2022.


\bibitem{debias} X. Wang, Z. Wu, L. Lian, and S. X. Yu, "Debiased learning from naturally imbalanced pseudo-labels," \emph{Proceedings of the IEEE/CVF Conference on Computer Vision and Pattern Recognition}, pp. 14647--14657, 2022.


\bibitem{iplc} G. Zhang, X. Qi, B. Yan, and G. Wang, "IPLC: iterative pseudo label correction guided by SAM for source-free domain adaptation in medical image segmentation," \emph{International Conference on Medical Image Computing and Computer-Assisted Intervention}, pp. 351--360, 2024.


\bibitem{information} J. Wu, H. Fan, Z. Li, G.-H. Liu, and S. Lin, "Information transfer in semi-supervised semantic segmentation," \emph{IEEE Transactions on Circuits and Systems for Video Technology}, vol. 34, no. 2, pp. 1174--1185, 2023.



\bibitem{cutout} T. DeVries and G. W. Taylor, "Improved regularization of convolutional neural networks with cutout," \emph{arXiv preprint arXiv}:1708.04552
        
        
        
        , 2017.

\bibitem{uamc} Y. Xia, D. Yang, Z. Yu, F. Liu, J. Cai, L. Yu, Z. Zhu, D. Xu, A. Yuille, and H. Roth, "Uncertainty-aware multi-view co-training for semi-supervised medical image segmentation and domain adaptation," \emph{Medical image analysis}, vol. 65, pp. 101766, 2020.



\bibitem{crmatch} Y. Fan, A. Kukleva, D. Dai, and B. Schiele, "Revisiting consistency regularization for semi-supervised learning," \emph{International Journal of Computer Vision}, vol. 131, no. 3, pp. 626--643, 2023.

\bibitem{crossmatch} B. Zhao, C. Wang, and S. Ding, "CrossMatch: Enhance Semi-Supervised Medical Image Segmentation with Perturbation Strategies and Knowledge Distillation," \emph{IEEE Journal of Biomedical and Health Informatics}, 2024.

\bibitem{edgeaware} Y. Yang, G. Sun, T. Zhang, R. Wang, and J. Su, "Semi-supervised medical image segmentation via weak-to-strong perturbation consistency and edge-aware contrastive representation," \emph{Medical Image Analysis}, vol. 101, pp. 103450, 2025.


\bibitem{cc3d} M. M. Hossam, A. E. Hassanien, and M. Shoman, "3D brain tumor segmentation scheme using K-mean clustering and connected component labeling algorithms," \emph{2010 10th International Conference on Intelligent Systems Design and Applications}, pp. 320--324, 2010.





\bibitem{ra3}
B. D. De Vos, J. M. Wolterink, P. A. de Jong, T. Leiner, M. A. Viergever, and I. Išgum, “ConvNet-based localization of anatomical structures in 3-D medical images,” \emph{IEEE Transactions on Medical Imaging}, vol. 36, no. 7, pp. 1470–1481, 2017. IEEE.








\bibitem{r1}
Y. Fan, A. Kukleva, D. Dai, and B. Schiele, “Revisiting consistency regularization for semi-supervised learning,” \emph{International Journal of Computer Vision}, vol. 131, no. 3, pp. 626–643, 2023. Springer.

\bibitem{mcf}
Y. Wang, B. Xiao, X. Bi, W. Li, and X. Gao, “MCF: Mutual correction framework for semi-supervised medical image segmentation,” in \emph{Proceedings of the IEEE/CVF Conference on Computer Vision and Pattern Recognition}, 2023, pp. 15651–15660.

\bibitem{mct}
Y. Wu, M. Xu, Z. Ge, J. Cai, and L. Zhang, “Semi-supervised left atrium segmentation with mutual consistency training,” in \emph{Medical Image Computing and Computer Assisted Intervention–MICCAI 2021: 24th International Conference, Strasbourg, France, September 27–October 1, 2021, Proceedings, Part II 24}, 2021, pp. 297–306. Springer.


\bibitem{r6}
X. Lu, L. Jiao, L. Li, F. Liu, X. Liu, S. Yang, Z. Feng, and P. Chen, “Weak-to-strong consistency learning for semi-supervised image segmentation,” \emph{IEEE Transactions on Geoscience and Remote Sensing}, vol. 61, pp. 1–15, 2023. IEEE.


\bibitem{noise}
Q. Xie, M.-T. Luong, E. Hovy, and Q. V. Le, “Self-training with noisy student improves ImageNet classification,” in \emph{Proceedings of the IEEE/CVF Conference on Computer Vision and Pattern Recognition}, 2020, pp. 10687–10698.


\bibitem{classmix}
V. Olsson, W. Tranheden, J. Pinto, and L. Svensson, “ClassMix: Segmentation-based data augmentation for semi-supervised learning,” in \emph{Proceedings of the IEEE/CVF Winter Conference on Applications of Computer Vision}, 2021, pp. 1369–1378.

\bibitem{cutmix}
S. Yun, D. Han, S. J. Oh, S. Chun, J. Choe, and Y. Yoo, “CutMix: Regularization strategy to train strong classifiers with localizable features,” in \emph{Proceedings of the IEEE/CVF International Conference on Computer Vision}, 2019, pp. 6023–6032.


\bibitem{imloss} Y. Guo, Y. Chen, L. Zhang, X. Liu, Y. Wang, X. Huang, and Z. Ma, "Im-loss: information maximization loss for spiking neural networks," \emph{Advances in Neural Information Processing Systems}, vol. 35, pp. 156--166, 2022.



\bibitem{rb1}
X. Wang, B. Zhang, L. Yu, and J. Xiao, "Hunting sparsity: Density-guided contrastive learning for semi-supervised semantic segmentation," in \emph{Proc. IEEE/CVF Conf. Comput. Vis. Pattern Recognit. (CVPR)}, pp. 3114–3123, 2023.

\bibitem{rb2}
H. Wu, X. Li, Y. Lin, and K.-T. Cheng, "Compete to win: Enhancing pseudo labels for barely-supervised medical image segmentation," \emph{IEEE Trans. Med. Imaging}, vol. 42, no. 11, pp. 3244–3255, 2023.



\bibitem{rb4}
P. Mi, J. Lin, Y. Zhou, Y. Shen, G. Luo, X. Sun, L. Cao, R. Fu, Q. Xu, and R. Ji, "Active teacher for semi-supervised object detection," in \emph{Proc. IEEE/CVF Conf. Comput. Vis. Pattern Recognit. (CVPR)}, pp. 14482–14491, 2022.

\bibitem{erdunet} H. Li, D.-H. Zhai, and Y. Xia, "ERDUnet: An efficient residual double-coding unet for medical image segmentation," \emph{IEEE Transactions on Circuits and Systems for Video Technology}, vol. 34, no. 4, pp. 2083--2096, 2023.




\bibitem{le1}
H. Basak and Z. Yin, “Pseudo-label guided contrastive learning for semi-supervised medical image segmentation,” in \emph{Proc. IEEE/CVF Conf. Comput. Vis. Pattern Recognit.}, pp. 19786–19797, 2023.


\bibitem{brats2021} U. Baid, S. Ghodasara, S. Mohan, M. Bilello, E. Calabrese, E. Colak, K. Farahani, J. Kalpathy-Cramer, F. C. Kitamura, S. Pati, et al., ``The RSNA-ASNR-MICCAI BRATS 2021 benchmark on brain tumor segmentation and radiogenomic classification,'' \emph{arXiv preprint arXiv}:2107.02314
        
        , 2021.






\bibitem{brats2020}
B. H. Menze, A. Jakab, S. Bauer, J. Kalpathy-Cramer, K. Farahani, J. Kirby, Y. Burren, N. Porz, J. Slotboom, R. Wiest, et al., “The multimodal brain tumor image segmentation benchmark (BRATS),” \emph{IEEE Trans. Med. Imaging}, vol. 34, no. 10, pp. 1993–2024, 2014.

\bibitem{IXI}
J. Chen, E. C. Frey, Y. He, W. P. Segars, Y. Li, and Y. Du, “TransMorph: Transformer for unsupervised medical image registration,” \emph{Med. Image Anal.}, 2022.


\bibitem{brats2023men}
D. LaBella, M. Adewole, M. Alonso-Basanta, T. Altes, S. M. Anwar, U. Baid, T. Bergquist, R. Bhalerao, S. Chen, V. Chung, G.-M. Conte, F. Dako, J. Eddy, I. Ezhov, D. Godfrey, F. Hilal, A. Familiar, K. Farahani, J. E. Iglesias, Z. Jiang, E. Johanson, A. F. Kazerooni, C. Kent, J. Kirkpatrick, F. Kofler, K. Van Leemput, H. B. Li, X. Liu, A. Mahtabfar, S. McBurney-Lin, R. McLean, Z. Meier, A. W. Moawad, J. Mongan, P. Nedelec, M. Pajot, M. Piraud, A. Rashid, Z. Reitman, R. T. Shinohara, Y. Velichko, C. Wang, P. Warman, W. Wiggins, M. Aboian, J. Albrecht, U. Anazodo, S. Bakas, A. Flanders, A. Janas, G. Khanna, M. G. Linguraru, B. Menze, A. Nada, A. M. Rauschecker, J. Rudie, N. H. Tahon, J. Villanueva-Meyer, B. Wiestler, and E. Calabrese, ``The ASNR-MICCAI Brain Tumor Segmentation (BraTS) Challenge 2023: Intracranial Meningioma,'' 2023.


\bibitem{uamt}
L. Yu, S. Wang, X. Li, C.-W. Fu, and P.-A. Heng, “Uncertainty-aware self-ensembling model for semi-supervised 3D left atrium segmentation,” in \emph{Medical Image Computing and Computer Assisted Intervention–MICCAI 2019: 22nd International Conference, Shenzhen, China, October 13–17, 2019, Proceedings, Part II 22}, 2019, pp. 605–613. Springer.

\bibitem{self} X. Lu, L. Jiao, L. Li, F. Liu, X. Liu, and S. Yang, "Self pseudo entropy knowledge distillation for semi-supervised semantic segmentation," \emph{IEEE Transactions on Circuits and Systems for Video Technology}, 2024.

\bibitem{fba}
Y. Chung, C. Lim, C. Huang, N. Marrouche, and J. Hamm, “FBA-Net: Foreground and background aware contrastive learning for semi-supervised atrium segmentation,” in \emph{Workshop on Medical Image Learning with Limited and Noisy Data}, 2023, pp. 106–116. Springer.

\bibitem{mctv1}
Y. Wu, Z. Ge, D. Zhang, M. Xu, L. Zhang, Y. Xia, and J. Cai, “Mutual consistency learning for semi-supervised medical image segmentation,” \emph{Medical Image Analysis}, vol. 81, p. 102530, 2022. Elsevier.

\end{thebibliography}
\end{document}